\documentclass[10pt,a4paper,authoryear]{nus-techreport}
\usepackage{float}
\graphicspath{{figures/}}

\usepackage{rethinking-opd}
\title{\texorpdfstring{\opdbalancedtitle{Rethinking On-Policy Distillation in Diffusion}{Models: \textcolor{reportSecondary}{Phenomenology and Insights}}}{Rethinking On-Policy Distillation in Diffusion Models: Phenomenology and Insights}}
\reportauthor[1]{Wei~Chow}
\reportauthor[1]{XiuYu~Zhang}
\reportauthor[2]{Yuxiang~Wei}
\reportauthor[8]{Bo~Chen}
\reportauthor[3]{Yingshuo~Wang}
\reportauthor[4]{Xian~Sun}
\reportauthor[5]{Yifei~Dong}
\reportauthor[1]{Xuan~Zhang}
\reportauthor[6]{Meng~Chu}
\reportauthor[1]{Pengyu~Li}
\reportauthor[10]{Xin~Li}
\reportauthor[1]{Dongyue~Lu}
\reportauthor[1]{Alan~Liang}
\reportauthor[7]{Ce~Zhang}
\reportauthor[9]{Feng~Cheng}
\reportauthor[10]{Jason~Li}
\reportauthor[1]{Lingdong~Kong}
\reportauthor[1,\dagger]{Leigang~Qu}
\reportauthor[1]{Tat-Seng~Chua}

\reportaffiliation[1]{National University of Singapore}
\reportaffiliation[2]{The Hong Kong Polytechnic University}
\reportaffiliation[3]{University of California, Berkeley}
\reportaffiliation[4]{Duke University}
\reportaffiliation[5]{University of Washington}
\reportaffiliation[6]{The Hong Kong University of Science and Technology}
\reportaffiliation[7]{Carnegie Mellon University}
\reportaffiliation[8]{Independent Research}
\reportaffiliation[9]{The University of North Carolina at Chapel Hill}
\reportaffiliation[10]{Nanyang Technological University}
\reportaffiliation[\ensuremath{\dagger}]{\thinspace Corresponding author}

\AtBeginDocument{\hypersetup{pdftitle={Rethinking On-Policy Distillation in Diffusion Models: Phenomenology and Insights}}}

\abstract{On-policy distillation combines diffusion specialists into one student by supervising the student's own denoising trajectories.
Yet its transfer mechanisms remain unclear, leaving open what transfers, what governs transfer, and when it is worthwhile.
We study existing on-policy distillation methods across text-to-image generation, image editing, and multi-image composition through budget-controlled comparisons, held-out evaluation, trajectory interventions, temporal supervision, teacher-source experiments, and direct joint reinforcement learning.
Transfer is selective, and neither teacher scores nor teacher resemblance consistently tracks student task gains.
Sparse methods lead at equal training time whereas dense methods lead at equal condition exposure, with query count per trajectory explaining $83.0\%$ of the equal-data lead.
Along trajectories, early high-noise steps already capture $80.2\%$ of the final edit direction in image editing, whereas $5.75\times$ wider late solver steps favor low-noise supervision in generation.
Narrow single-route reinforcement learning without cross-route regularization increases teacher target scores by up to $28.2\%$ yet reduces student held-out editing by up to $42.5\%$, whereas joint reinforcement learning preserves a broader task profile with online rewards.
Grounded in these mechanisms, we propose PrefixOPD, an efficient on-policy distillation method that aligns supervision with task-specific trajectory dynamics.
PrefixOPD pairs low-noise generation queries with step-$1$ prefix editing rollouts and initialization-referenced target extrapolation, skipping $81.8\%$ of editing forwards while preserving the full-rollout stop-gradient update.
Over full-rollout distillation, PrefixOPD reduces active training time by $44.5\%$ while improving text-to-image generation by $3.7\%$, overall eight-route editing by $9.3\%$, and the three-route editing diagnostic suite by $15.5\%$, and outperforms equal-compute baselines by up to $17.4\%$ across backbones and mixtures, surpassing equal-data extrapolation at nearly half the cost.
We hope these findings and PrefixOPD inspire efficient multi-teacher post-training for generative models.}

\begin{document}

\maketitle

\reportabstract

\begin{figure}[H]
\centering
\includegraphics[width=\textwidth]{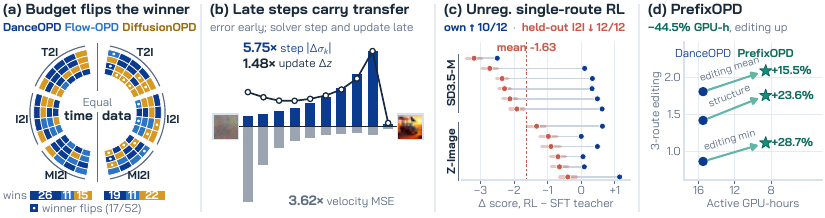}
\vspace{-0.38cm}
\caption{\textbf{Overview of diffusion on-policy distillation.} (a) Changing the budget reverses method rankings. (b) Wider late solver steps amplify low-noise updates. (c) Narrow single-route reinforcement teachers degrade held-out editing, whereas (d) PrefixOPD improves editing at lower cost.}
\label{fig:teaser}
\vspace{-0.15cm}
\end{figure}

\section{Introduction}

Diffusion models increasingly support text-to-image generation, image editing, and multi-image composition together~\citep{ho2020denoising,rombach2022high,xiao2025omnigen}, including instruction-guided and iterative editing~\citep{brooks2023instructpix2pix,chow2026editmgt,chow2026weave}.
However, supervised fine-tuning (SFT) and reinforcement learning (RL) improve individual routes while degrading others, motivating multi-teacher distillation into a shared student~\citep{hinton2015distilling,agarwal2024policy}.
Diffusion on-policy distillation (OPD) supervises student trajectories by matching sparse velocity predictions~\citep{zhou2026danceopd}, dense transition means~\citep{li2026diffusionopd}, or regularized stochastic transitions~\citep{fang2026flow}.

\begin{figure}[t]
\centering
\vspace{-0.3cm}
\includegraphics[width=\textwidth]{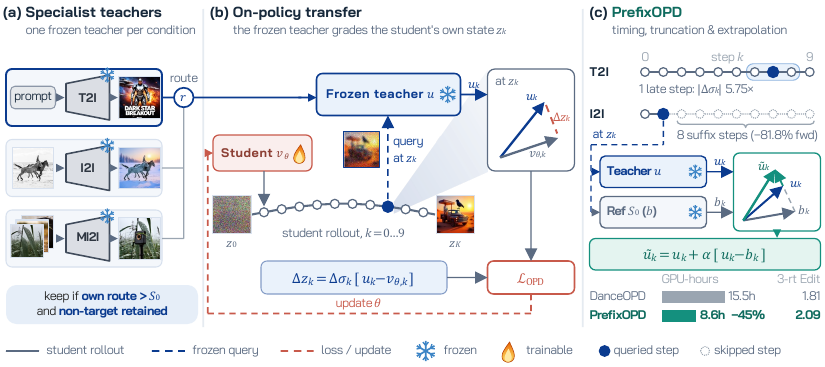}
\vspace{-0.42cm}
\caption{\textbf{Overview of multi-teacher distillation and PrefixOPD.} (a) Teachers are filtered by target gain and route preservation, (b) routed teachers supervise student states, and (c) PrefixOPD combines asymmetric timing, prefix rollout, and target extrapolation.}
\label{fig:study-framework}
\vspace{-0.15cm}
\end{figure}

However, what diffusion on-policy distillation transfers and what governs student gains remain largely unexplored.
Specifically, a specialist's own score does not determine what a shared student learns, and teacher resemblance need not imply stronger task behavior~\citep{tiapkin2025teacherhacking}.
Prior studies analyze capacity gaps in static distillation~\citep{cho2019efficacy,mirzadeh2020improved,busbridge2025distillation,li2025small} and student-relative evaluation under controlled cost in language models~\citep{li2026rethinking,xie2026multiteacher}.
In diffusion models, trajectory supervision further interacts with continuous noise schedules, numerical ODE solvers, and clean source conditioning, yet standard multi-teacher pipelines typically query generation and editing symmetrically across full rollouts---and even phase-selective skipping~\citep{yan2026phaseopd} does not align query timing with the opposing trajectory dynamics of generation versus editing---while selecting teachers by target score alone.
Accordingly, we study what on-policy distillation transfers, what determines transfer at student-visited states, and when distillation complements joint reinforcement learning in Figures~\ref{fig:teaser} and~\ref{fig:study-framework}.

Experiments across backbones and mixtures show that on-policy distillation transfers specialist strengths selectively rather than uniformly.
First, sparse methods lead at equal time whereas dense methods lead at equal condition exposure in Figure~\ref{fig:teaser}(a), with query count driving $83.0\%$ of the equal-data lead.
Internally, representation changes remain localized in middle and late decoding blocks, while students exceed routed teachers in $72.2\%$ of held-out comparisons.
Second, transfer depends on query timing and teacher choice, as early high-noise steps capture $80.2\%$ of the final edit direction in image editing whereas $5.75\times$ wider late solver steps favor low-noise supervision in generation in Figure~\ref{fig:teaser}(b).
Meanwhile, narrow single-route reinforcement learning without cross-route regularization increases teacher target scores by up to $28.2\%$ yet reduces student held-out editing by up to $42.5\%$ once supervised teachers exceed initialization in Figure~\ref{fig:teaser}(c).
Third, direct joint reinforcement learning preserves a broader profile with online rewards, whereas distillation excels at modular recovery from frozen teachers.

Grounded in these findings, PrefixOPD aligns on-policy distillation with asymmetric generation and editing dynamics.
Specifically, PrefixOPD keeps generation queries at late low-noise steps where wider solver intervals amplify updates, and places editing queries at the first step after pure noise where edit direction forms.
Because causal detached velocity matching with suffix-independent queries depends only on states up to the last query, PrefixOPD truncates each rollout there, skipping $81.8\%$ of editing forwards while exactly preserving the full-rollout stop-gradient update.
It adds $25\%$ initialization-referenced target extrapolation on early editing queries while leaving generation unchanged.

Across generation, editing, and composition benchmarks in Figure~\ref{fig:teaser}(d), PrefixOPD reduces active training time by $44.5\%$ over existing full-rollout distillation while improving text-to-image generation by $3.7\%$, overall eight-route editing by $9.3\%$, and the three-route editing diagnostic suite by $15.5\%$.
PrefixOPD already surpasses full-budget baseline editing by $9.4\%$ in roughly a quarter of the training time, and at equal compute it outperforms existing baselines by up to $17.4\%$ on editing and $8.4\%$ on generation while surpassing equal-data full-rollout extrapolation at nearly half the cost.
We hope our mechanistic findings and PrefixOPD inspire efficient multi-teacher post-training across generative models.

We summarize our main contributions below.
\begin{itemize}
\item \textbf{Method.} We propose PrefixOPD, an asymmetric on-policy distillation method coupling task-specific query timing, prefix rollout truncation, and initialization-referenced target extrapolation.
\item \textbf{Results.} PrefixOPD reduces active training time by $44.5\%$ over existing full-rollout distillation while improving generation by $3.7\%$, overall eight-route editing by $9.3\%$, and the three-route editing diagnostic suite by $15.5\%$, exceeding equal-compute baselines by up to $17.4\%$.
\item \textbf{Analysis.} Controlled experiments clarify what on-policy distillation transfers, how query timing and teacher choice govern student gains, and how distillation complements joint reinforcement learning.
\end{itemize}

\section{Problem Formulation}
\label{sec:setup}

This section formulates multi-teacher on-policy distillation: Section~\ref{sec:notation} defines routed trajectories, Section~\ref{sec:opd-objectives} specifies the distillation objectives, target extrapolation, and prefix truncation, and Section~\ref{sec:time-conventions} states our evaluation, noise-schedule, and diagnostic conventions.

\subsection{Notation}
\label{sec:notation}

Let $x=(p,I^{\mathrm{src}},\mathbf I^{\mathrm{ref}})\sim\mathcal D$ denote a condition, where $p$ is an instruction, $I^{\mathrm{src}}$ is an optional source image, and $\mathbf I^{\mathrm{ref}}$ is an ordered tuple of labeled reference images.
Missing image inputs are represented by $\emptyset$, and $\mathcal R$ denotes the set of specialist training routes.
For each trajectory, the routing rule $r(x)\in\mathcal R$ assigns one frozen teacher $T_{r(x)}$ acquired through supervised fine-tuning or reinforcement learning~\citep{zheng2026diffusionnft,black2024training,prabhudesai2023aligning}.

Let $S_\theta$ be the student with trainable parameters $\theta$ and fixed initialization $S_0$.
Under method $m$, the current student generates a denoising trajectory
\begin{equation}
\tau=(z_0,\ldots,z_K)
\sim p_\theta^{(m)}(\cdot\mid x),
\label{eq:rollout}
\end{equation}
across decreasing noise levels $\sigma_0>\cdots>\sigma_K$, where $z_k$ stores the target latent and any solver history required by method $m$, and $p_\theta^{(m)}$ captures initial Gaussian noise and sampler stochasticity.
At each visited state $z_k$ for $k\in\{0,\ldots,K-1\}$, student and teacher velocity predictions are
\begin{equation}
v_{\theta,k}
\coloneqq v_\theta(z_k,\sigma_k,x),
\qquad
u_k
\coloneqq v_{T_{r(x)}}(z_k,\sigma_k,x),
\label{eq:velocities}
\end{equation}
where $r$ abbreviates $r(x)$.
Both predictions act on the same target latent at state $z_k$.

\subsection{On-Policy Distillation}
\label{sec:opd-objectives}

Method $m$ queries $T_r$ at step indices $\mathcal I_m(\tau,x)\subseteq\{0,\ldots,K-1\}$ and minimizes
\begin{equation}
\mathcal L_{\mathrm{OPD}}^{(m)}(\theta)
=
\mathbb E_{\substack{
x\sim\mathcal D\\
\tau\sim p_\theta^{(m)}(\cdot\mid x)
}}
\left[
\ell_m\!\left(
\theta;\tau,x,
\{(k,u_k)\mid k\in\mathcal I_m(\tau,x)\}
\right)
\right],
\label{eq:opd-general}
\end{equation}
where random query selection enters the expectation and $\ell_m$ keeps native weighting and regularization.
All methods supervise states visited by the current student~\citep{ross2011reduction}.

Three representative methods instantiate Equation~\ref{eq:opd-general} differently (abbreviating its expectation and query set by method).
Specifically, DanceOPD matches velocities via $\operatorname{MSE}(v_{\theta,k},u_k)\coloneqq D^{-1}\lVert v_{\theta,k}-u_k\rVert_2^2$ over $D$-dimensional target latents at sparse states~\citep{zhou2026danceopd} as
\begin{equation}
\mathcal L_{\mathrm{OPD}}^{(\mathrm{Dance})}(\theta)
=
\mathbb E_{\mathrm{Dance}}
\left[
\sum_{k\in\mathcal I_{\mathrm{Dance}}}
a_k\,
\operatorname{MSE}(v_{\theta,k},u_k)
\right],
\label{eq:danceopd}
\end{equation}
with native query weights $a_k\geq 0$.
In contrast, DiffusionOPD matches transition means $\mu_{\theta,k}$ and $\mu_{T_{r(x)},k}$ across dense queries with weights $\omega_k\geq 0$~\citep{li2026diffusionopd}, formulated as
\begin{equation}
\mathcal L_{\mathrm{OPD}}^{(\mathrm{Diffusion})}(\theta)
=
\mathbb E_{\mathrm{Diffusion}}
\left[
\sum_{k\in\mathcal I_{\mathrm{Diffusion}}}
\omega_k
\left\|
\mu_{\theta,k}(z_k,x)
-
\mu_{T_{r(x)},k}(z_k,x)
\right\|_2^2
\right].
\label{eq:diffusionopd}
\end{equation}
Meanwhile, Flow-OPD combines stochastic student transitions with clipped policy ratios and manifold-anchoring regularization toward $S_0$~\citep{fang2026flow}.

\noindent\textbf{Target extrapolation and prefix rollout truncation.}
At state $z_k$, let $B_r$ be a frozen reference for route $r$ with velocity $b_{r,k}\coloneqq v_{B_r}(z_k,\sigma_k,x)$, writing $u_{r,k}\coloneqq u_k$ (shown as $b_k,u_k,\widetilde u_k$ in Figure~\ref{fig:study-framework}(c)).
Under target extrapolation~\citep{yang2026gopd,lin2026dreopd,wei2026stepopd}, the transformed target is
\begin{equation}
\widetilde u_{r,k}(\lambda_r)
=(1-\lambda_r)b_{r,k}+\lambda_r u_{r,k}
=u_{r,k}+\alpha_r(u_{r,k}-b_{r,k}),\qquad \alpha_r=\lambda_r-1,
\label{eq:shared-extrapolation}
\end{equation}
where $0<\lambda_r<1$ interpolates toward $B_r$, $\lambda_r=1$ matches the teacher at $\alpha_r=0$, and $\lambda_r>1$ extrapolates $u_{r,k}-b_{r,k}$.
For a causal rollout solver whose query set $\mathcal I_r\subseteq\{0,\ldots,K-1\}$ and targets $\widetilde u_{r,k}$ are independent of suffix states $k>k_{\max}=\max\mathcal I_r$, truncating rollouts after $k_{\max}$ omits unused suffix forwards under
\begin{equation}
\mathcal L_{\mathrm{vel}}(\theta)=
\mathbb E_{\substack{x\sim\mathcal D\\ \tau_{\leq k_{\max}}\sim p_\theta(\cdot\mid x)}}\!\left[
\frac{1}{|\mathcal I_r|}
\sum_{k\in\mathcal I_r}
\operatorname{MSE}\!\left(
v_\theta(\operatorname{sg}(z_k),\sigma_k,x),
\operatorname{sg}(\widetilde u_{r,k})
\right)\right],
\label{eq:clx-loss}
\end{equation}
where $\operatorname{sg}(\cdot)$ holds visited states, targets $\widetilde u_{r,k}\coloneqq\widetilde u_{r,k}(\lambda_r)$, and $p_\theta$ fixed during differentiation; truncation thus exactly preserves the full-rollout stop-gradient update without differentiating through $p_\theta$.

\subsection{Evaluation And Time Conventions}
\label{sec:time-conventions}

A teacher's score on its own outputs measures endpoint behavior rather than usefulness to the student.
Therefore, we evaluate transfer relative to $S_0$ across $13$ routes covering $5$ text-to-image (T2I), $4$ image editing (I2I), and $4$ multi-image composition (MI2I) routes, including held-out I2I and MI2I suites.
We further validate our primary evaluator against an open-source vision-language judge in Appendix~\ref{app:extension-judging}, treating gains over initialization, gains over the teacher, and teacher resemblance as distinct quantities.

We distinguish noise level $\sigma$ from denoising progress $\rho_k$ under $z_\sigma=(1-\sigma)z_{\mathrm{clean}}+\sigma\epsilon$ for $\sigma\in[0,1]$, clean target latent $z_{\mathrm{clean}}$, and Gaussian noise $\epsilon$.
Along a $K$-step rollout with intervals $\Delta\sigma_k\coloneqq\sigma_{k+1}-\sigma_k$, denoising progress is $\rho_k\coloneqq k/K$.
Evaluating $\sigma(\eta)=s\eta/(1+(s-1)\eta)$ at $\eta_k=1-\rho_k$ with shift $s\geq 1$ advances from $\sigma_0=1$ to $\sigma_K=0$.
Consequently, early high-noise steps have small $\rho_k$, whereas late low-noise steps have wider $|\Delta\sigma_k|$ when $s>1$ (with a small terminal residual at the final step on SD3.5-M, yielding $5.75\times$ wider active intervals from step $k=1$ to $k=8$ at $s=3$).
An inference intervention alters one fixed trajectory, whereas a training query updates the student for subsequent trajectories.

For one solver step $\Phi_k(z_k,v)$ under velocity $v$, replacing $z_{k+1}=\Phi_k(z_k,v_{\theta,k})$ with $u_k$ injects the correction $\Delta z_k\coloneqq\Phi_k(z_k,u_k)-\Phi_k(z_k,v_{\theta,k})$ with norm $\lVert\Delta z_k\rVert_{\mathrm{RMS}}\coloneqq D^{-1/2}\lVert\Delta z_k\rVert_2$, and velocity disagreement is $d_k\coloneqq\lVert v_{\theta,k}-u_k\rVert_2$.
We measure teacher resemblance via $\Delta\mathrm{CKA}\coloneqq\mathrm{CKA}(S_\theta,T_r)-\mathrm{CKA}(S_0,T_r)$~\citep{kornblith2019similarity}, and measure patch direction alignment on decoded one-step clean images $\hat I_k$ via $\cos(\Delta\phi_k,\Delta\phi_{K-1})$ with source-subtracted DINOv2 patch feature shifts $\Delta\phi_k\coloneqq\phi(\hat I_k)-\phi(I^{\mathrm{src}})$~\citep{oquab2023dinov2}.
Unlike progressive step reduction~\citep{salimans2022progressive,song2023consistency}, we hold inference schedules fixed, with related work and protocols in Appendices~\ref{sec:related-work}, \ref{app:opd-common-design}, and~\ref{app:teacher-evaluation}.

\section{What Does On-Policy Distillation Transfer?}
\label{sec:consolidation}

\begin{table}[t]
\centering
\vspace{-0.5cm}
\includegraphics[width=\textwidth]{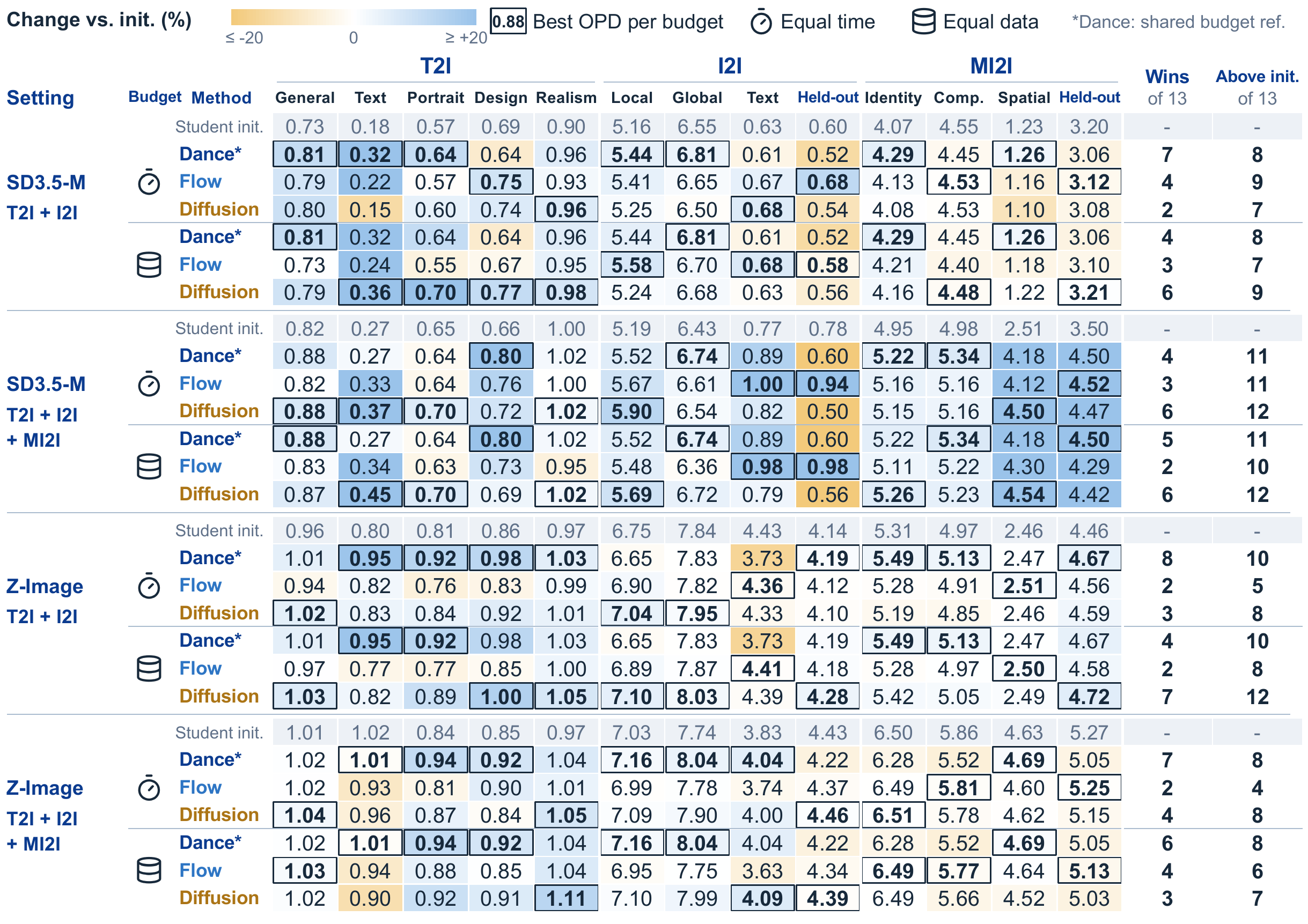}
\vspace{-0.5cm}
\caption{\textbf{Comparison of on-policy distillation methods across routes and backbones.} Matching condition exposure rather than training time reverses route winners and shifts the lead from DanceOPD to DiffusionOPD, with DanceOPD anchoring both budgets as the shared reference (two-decimal ties broken by unrounded scores).}
\vspace{-0.15cm}
\label{tab:route-budget-map}
\end{table}

The first question is what on-policy distillation transfers across routes and interfaces.
We first compare student task profiles under equal-time and equal-data budgets in Section~\ref{sec:budget-consolidation}.
After that, Section~\ref{sec:heldout-behavior} separates held-out task gains from teacher resemblance.

\subsection{Budget-Dependent Transfer}
\label{sec:budget-consolidation}

We audit paired supervised fine-tuning and reinforcement learning teachers in Appendix~\ref{app:teacher-performance}, and then compare DanceOPD, DiffusionOPD, and Flow-OPD on two mixtures across SD3.5-M and Z-Image~\citep{zhou2026danceopd,li2026diffusionopd,fang2026flow}.
Specifically, TI combines general T2I and general I2I routes, whereas TIM adds MI2I composition.
Equal-time runs match elapsed training time and equal-data runs match condition exposure under shared protocols in Appendix~\ref{app:opd-common-design} and Table~\ref{tab:route-budget-map}.

\noindent\textbf{Specialist performance does not establish transfer.}
Evaluating multi-teacher distillation requires measuring the full student profile rather than teacher target scores.
In Table~\ref{tab:teacher-performance}, narrow single-route reinforcement learning improves the optimized route in $72.7\%$ of paired comparisons ($77.3\%$ non-decreasing at two-decimal precision), yet nearly every specialist degrades other routes when deployed alone.

\noindent\textbf{The training budget changes the winning method via query count.}
Matching condition exposure rather than elapsed training time reverses the leading distillation family across mixtures.
Specifically, DanceOPD wins the most routes in $3$ of $4$ equal-time settings, whereas DiffusionOPD leads in $3$ of $4$ equal-data settings, changing winners in $32.7\%$ of matched route comparisons in Figure~\ref{fig:budget-transitions}(a).
Moreover, matching condition exposure increases DiffusionOPD's mean gain over initialization by up to $11.8$ percentage points, improving T2I Mean by $10.7\%$ on SD3.5-M TI and eight-route Editing Mean by $2.4\%$ on Z-Image TI.
Crossing $1$ versus $4$ queries per trajectory with velocity versus transition targets in Figure~\ref{fig:budget-transitions}(b) isolates query count.
Increasing velocity queries halves updates per hour and reduces equal-time gain by $80.0\%$, yet increases equal-data gain to $10.9\%$ and accounts for $83.0\%$ of DiffusionOPD's lead.
Conversely, reducing transition matching to $1$ query restores a $6.5\%$ equal-time gain while reducing equal-data gain by $41.9\%$.

\noindent\textbf{On-policy distillation transfers specialist gains selectively.}
Distilled students acquire specialist strengths selectively, shaped by initialization headroom and schedule curvature.
In Table~\ref{tab:route-budget-map}, DiffusionOPD improves $12$ of $13$ routes over initialization on SD3.5-M TIM at both budgets and lifts held-out I2I by $3.4\%$ and held-out MI2I by $5.8\%$ on Z-Image TI at equal data.
Meanwhile, on SD3.5-M TIM, DanceOPD improves trained MI2I composition by $7.2\%$ and held-out MI2I by $28.6\%$, while Flow-OPD improves held-out MI2I by $29.1\%$ under equal time.
In contrast, Z-Image starts nearly $7\times$ higher on held-out I2I, leaving less headroom where DanceOPD leads TIM wins and Flow-OPD's manifold regularization preserves multi-image routes at equal data, whereas SD3.5-M's steeper schedule favors dense supervision.

\subsection{Held-Out Behavior Beyond Teacher Resemblance}
\label{sec:heldout-behavior}

\noindent\textbf{Students often exceed routed teachers on combined held-out constraints.}
Evaluating each image's weakest rubric criterion highlights consistency across multiple constraints on held-out tasks.
In Appendix~\ref{app:complete-heldout}, students exceed their routed teachers in $72.2\%$ of held-out comparisons under the per-image minimum score, reversing the held-out ordering between DanceOPD and DiffusionOPD.

\noindent\textbf{Teacher representation alignment concentrates in middle and late blocks.}
Representation changes remain localized rather than copying teacher features across all layers.
For TI, initialization is itself reused as the editing teacher to anchor editing preservation while distilling T2I; therefore, CKA relative to initialization directly tracks representation preservation.
In contrast, for TIM, $30.6\%$ of CKA changes are positive and concentrate in deeper decoding blocks (most prominently at early high-noise step $1$ in Figure~\ref{fig:cka-depth-ranges}), whereas raw feature discrepancy peaks broadly at early steps and deep blocks~\citep{lin2026focusopd}.

\noindent\textbf{Patch correspondence does not reproduce held-out task rankings.}
Static patch similarity fails to track task rankings across distilled students.
Across held-out evaluations, static DINOv2 correspondence varies narrowly across methods and disagrees with held-out rubric scores because unedited backgrounds dominate patch matching, motivating our trajectory analysis next.

\section{What Determines Transfer?}
\label{sec:mechanism}

\begin{figure}[t]
\centering
\includegraphics[width=\textwidth]{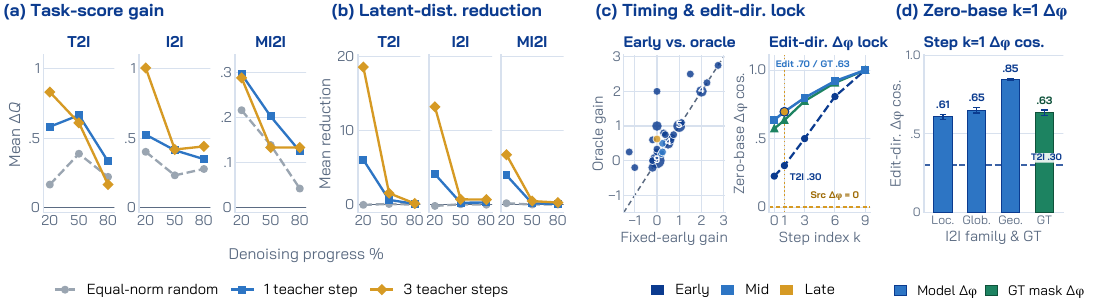}
\vspace{-0.52cm}
\caption{\textbf{Trajectory interventions and edit direction.} Teacher steps exceed random controls on (a) task gain and (b) latent distance, while (c) early steps match the oracle and lock edit direction (cosine $0.70$ to final shift, $0.63$ in masks, and $80.2\%$ of final ground-truth alignment at step $1$) across (d) edit families.}
\label{fig:trajectory-interventions}
\label{fig:timing-selection}
\end{figure}

The second question is what determines transfer at states visited by the student.
We first test correction direction, dose, and edit direction on frozen trajectories in Section~\ref{sec:frozen-trajectory}.
Section~\ref{sec:teacher-strength} then analyzes temporal supervision and solver step sizes, and Section~\ref{sec:teacher-source} examines paired teacher sources.

\subsection{Correction And Continuation On Frozen Trajectories}
\label{sec:frozen-trajectory}

Following Section~\ref{sec:time-conventions}, we inject the routed teacher prediction at $20\%$, $50\%$, or $80\%$ denoising progress and continue all later steps with the student.
At each stage, we compare random controls of equal norm, $1$ teacher step, and windows of $3$ teacher steps in Appendix~\ref{app:trajectory-interventions}.

\noindent\textbf{Correction direction matters.}
Directed single teacher steps yield larger mean task gains than random controls of equal norm across all three task families.
In Figure~\ref{fig:trajectory-interventions}(a), single teacher steps double the T2I gain of random controls ($2.04\times$) and increase mean gains by $41.0\%$ on I2I and $57.3\%$ on MI2I.
Moreover, teacher corrections consistently move final latents closer to the teacher trajectory in Figure~\ref{fig:trajectory-interventions}(b), whereas random controls yield nearly zero reduction.

\noindent\textbf{Early steps at high noise establish target edit direction in image editing.}
Early interventions produce the largest gains, with $2.28\times$ higher I2I gains at $20\%$ than at $80\%$ progress under three-step windows in Figure~\ref{fig:trajectory-interventions}(a) (and up to $3.45\times$ on the expanded $32$-condition I2I audit in Appendix~\ref{app:trajectory-interventions}).
Specifically, widening an early intervention from $1$ to $3$ steps increases early I2I gain by $91.3\%$ in Figure~\ref{fig:trajectory-interventions}(a) ($44.8\%$ across all stages, and $36.4\%$ on the expanded I2I audit) while leaving overall T2I mean within $1.8\%$, as local and text edits peak under $1$ early step whereas global and geometric edits benefit from wider windows.
In Figures~\ref{fig:trajectory-interventions}(c) and~\ref{fig:trajectory-interventions}(d), DINOv2 direction similarity on edited I2I patches reaches $69.6\%$ after $1$ step, exceeding T2I by $2.30\times$.
Inside external ground-truth masks, trajectory alignment with the final masked prediction reaches $63.4\%$ after $1$ step in Figure~\ref{fig:trajectory-interventions}(c), while alignment with the external ground-truth shift $\Delta\bar\phi^{\mathrm{gt}}$ (Appendix~\ref{app:absorption-diagnostics}) reaches $80.2\%$ of its final value at step $1$ and $91.5\%$ by step $3$.

\noindent\textbf{Fixed early intervention matches disagreement selection.}
Choosing the state with maximum velocity disagreement selects the early position in $91.7\%$ of cases and matches the best timing in $68.8\%$, identical to fixed early placement in Figure~\ref{fig:trajectory-interventions}(c).
Additionally, fixed early placement matches disagreement selection across task families and slightly exceeds it in overall mean gain ($0.490$ versus $0.486$), confirming that fixed early queries capture the benefit of adaptive timing selection without extra teacher evaluations.

\subsection{Temporal Supervision And Solver Step Amplification}
\label{sec:teacher-strength}

Whereas an intervention changes one fixed trajectory, training updates the student velocity field across AdamW steps.
Under fixed SD3.5-M initialization, teachers, and data, we compare low, middle, high, and uniform noise schedules alongside step-damped and uniform-step solver controls.

\begin{figure}[t]
\centering
\includegraphics[width=\textwidth]{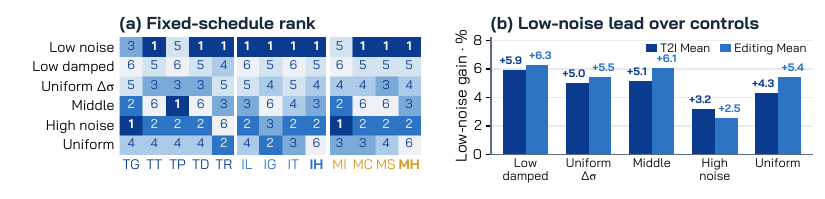}
\vspace{-0.92cm}
\caption{\textbf{Temporal supervision across benchmark routes.} (a) Low-noise supervision leads fixed schedules on $10$ of $13$ routes (T2I: TG/TT/TP/TD/TR; I2I: IL/IG/IT/IH; MI2I: MI/MC/MS/MH for General/Text/Portrait/Design/Realism, Local/Global/Text/Held-out, and Identity/Comp./Spatial/Held-out) whereas solver-damped controls win none, and (b) low-noise supervision consistently outperforms alternative schedules and solver controls on both T2I Mean and eight-route Editing Mean.}
\label{fig:temporal-supervision-results}
\end{figure}

\noindent\textbf{Supervision at low noise leads symmetric training across routes.}
Standard supervision at low noise ranks first on $10$ of $13$ routes in Figure~\ref{fig:temporal-supervision-results}(a) and Table~\ref{tab:temporal-scheduling}, exceeding alternative schedules and solver controls by up to $5.9\%$ on T2I Mean ($5.8\%$ at two-decimal table precision) and $6.3\%$ on eight-route Editing Mean in Figure~\ref{fig:temporal-supervision-results}(b).
In contrast, removing late step expansion eliminates all $10$ wins, reducing route averages by $5.6\%$ to $5.9\%$ under damped weighting and $4.8\%$ to $5.2\%$ under uniform solver steps.

\begin{figure}[t]
\centering
\includegraphics[width=\textwidth]{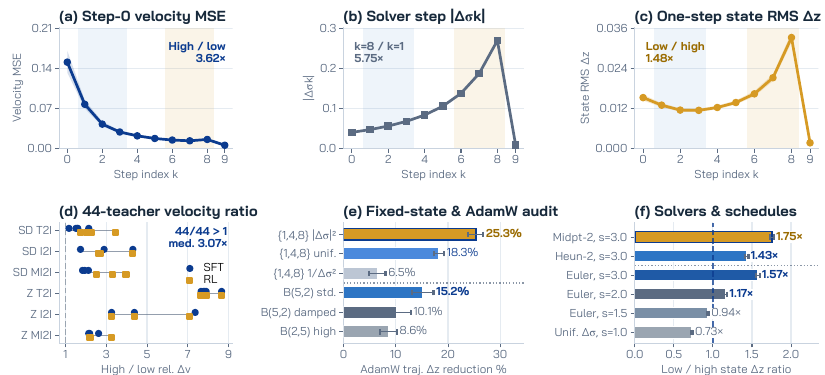}
\vspace{-0.42cm}
\caption{\textbf{Solver step amplification and schedule generality.} Despite higher early error in (a) rollouts and (d) teachers, (b) wider late steps yield (c) larger updates and (e) stronger AdamW error reduction under low-noise $\mathrm{Beta}(5,2)$ versus high-noise $\mathrm{Beta}(2,5)$ sampling, holding across (f) shifted solvers and schedules.}
\label{fig:e4-solver-probe}
\end{figure}

\noindent\textbf{Wider late solver steps amplify updates at low noise under shifted schedules.}
Initial velocity MSE is $3.62\times$ higher at high noise and early relative velocity disagreement is $3.07\times$ higher median in Figures~\ref{fig:e4-solver-probe}(a) and~\ref{fig:e4-solver-probe}(d).
Nevertheless, the $5.75\times$ wider late solver step in Figure~\ref{fig:e4-solver-probe}(b) yields $1.48\times$ larger DPM-2 state updates in Figure~\ref{fig:e4-solver-probe}(c).
Across midpoint, Euler, and Heun-2 solvers in Figure~\ref{fig:e4-solver-probe}(f) (alongside the $1.48\times$ rollout ratio in Figure~\ref{fig:e4-solver-probe}(c)), shifted queries at low noise yield $1.43\times$ to $1.75\times$ larger state updates than at high noise.
Conversely, removing the schedule shift ($s=1.0$) reduces the Euler update ratio by $53.5\%$ to $0.73\times$, flipping update magnitude toward high noise (consistent with the $1.9\%$ unshifted T2I timing reversal in Table~\ref{tab:clx-all-ablations}).
In Figure~\ref{fig:e4-solver-probe}(e), damped weighting rotates AdamW updates away from late steps and reduces trajectory error reduction by $33.6\%$, while squared step weighting on fixed states yields $3.89\times$ the error reduction of inverse weighting.
Together, early edit formation in I2I and late step amplification in shifted T2I generation motivate keeping T2I queries at low noise while moving I2I queries to step $1$.

\subsection{Teacher Transferability}
\label{sec:teacher-source}

\begin{figure}[t]
\centering
\includegraphics[width=\textwidth]{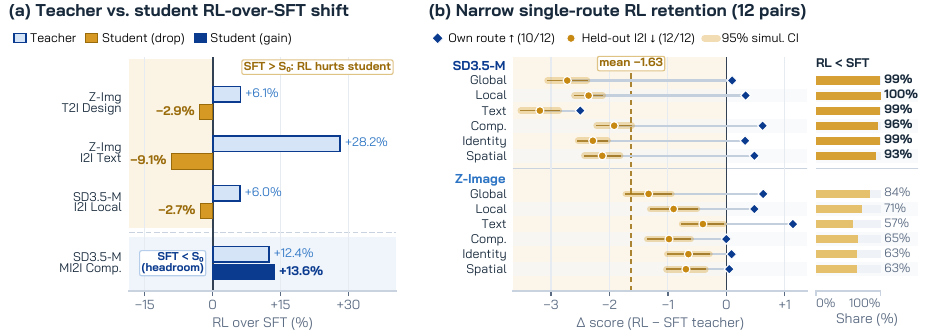}
\vspace{-0.12cm}
\caption{\textbf{Teacher-source transfer and cross-route retention.} (a) Teacher and student rankings diverge across four SFT-versus-RL pairs; (b) narrow single-route RL teachers improve $10$ of $12$ target routes yet degrade held-out editing in all $12$ pairs (right bars show held-out shares with RL below SFT).}
\label{fig:teacher-headroom-cases}
\label{fig:teacher-retention-12pairs}
\end{figure}

We next hold DanceOPD training fixed while swapping supervised fine-tuning and reinforcement learning teachers on Z-Image T2I design, Z-Image I2I text editing, SD3.5-M I2I local editing, and SD3.5-M MI2I composition in Figure~\ref{fig:teacher-headroom-cases}(a) and Table~\ref{tab:teacher-strength}.

\noindent\textbf{A stronger teacher can produce a weaker student.}
Once a supervised teacher exceeds student initialization, narrow single-route reinforcement learning without cross-route regularization degrades preservation across routes and weakens the student.
In T2I design, the reinforcement teacher scores $6.1\%$ higher on its own route yet yields a student $2.9\%$ below the student of the supervised teacher.
Similarly, in Z-Image text editing, reinforcement learning increases teacher target score by $28.2\%$ while reducing held-out editing by $8.9\%$, leaving its student $9.1\%$ lower on text editing and $4.4\%$ lower on held-out editing.
On SD3.5-M local editing, where supervised fine-tuning exceeds initialization by $3.8\%$, reinforcement learning increases teacher local editing by $6.0\%$ while reducing text editing by $65.9\%$ and held-out I2I by $84.9\%$.
Consequently, its student scores $2.7\%$ lower on local editing and $42.5\%$ lower on held-out I2I.

\noindent\textbf{The preferred teacher reverses when supervised fine-tuning lacks headroom.}
In SD3.5-M MI2I composition, the supervised teacher falls $5.7\%$ below initialization in Table~\ref{tab:teacher-strength}.
Reinforcement learning restores a $6.0\%$ composition lead and a $27.1\%$ held-out MI2I lead over initialization ($+12.4\%$ over the supervised teacher in Figure~\ref{fig:teacher-headroom-cases}(a)).
Therefore, its student outperforms the student of the supervised teacher by $13.6\%$ on composition and $47.5\%$ on held-out MI2I.

\noindent\textbf{Target improvement and preservation across routes guide teacher selection.}
Across all $12$ paired editing specialists in Figure~\ref{fig:teacher-retention-12pairs}(b), narrow single-route reinforcement learning improves $83.3\%$ of target routes yet reduces held-out I2I in $100\%$ of pairs, while velocity disagreement fails to separate good and bad teachers.
These results support two validation checks before multi-teacher distillation.
\begin{enumerate}
\item \textbf{Rule 1 on target improvement.} A candidate teacher should exceed student initialization on its target validation split, motivating reinforcement or joint training when supervised fine-tuning falls short.
\item \textbf{Rule 2 on preservation across routes.} Among candidates meeting Rule 1, checkpoints preserving other routes transfer more reliably than narrow specialists across Tables~\ref{tab:teacher-strength} and~\ref{tab:clx-all-ablations}.
\end{enumerate}

\section{When Is Distillation Worthwhile And How Can It Be Improved?}
\label{sec:evaluation}
\label{sec:joint-optimization}

Sections~\ref{sec:within-interface-reward} and~\ref{sec:cross-interface-consolidation} examine when distillation is worthwhile relative to parameter merging and joint reinforcement learning, and Section~\ref{sec:clx-evaluation} introduces PrefixOPD.

\subsection{Within-Interface Reward Transfer}
\label{sec:within-interface-reward}

Starting from shared general T2I initialization, we compare PickScore, CLIPScore, and HPSv2.1 specialists~\citep{kirstain2023pick,hessel2021clipscore,wu2023human} across single-teacher distillation, joint distillation, and direct joint reinforcement learning.
Specifically, single-teacher DanceOPD and DiffusionOPD transfer positive gains on each optimized target reward.
However, standalone HPSv2.1 tuning reduces general T2I quality by $30.6\%$ ($31.3\%$ at two-decimal precision).
In contrast, joint DiffusionOPD across all $3$ teachers limits this drop to $9.6\%$ and exceeds direct joint reinforcement learning by $8.2\%$ on general T2I quality ($8.7\%$ at two-decimal precision).

\subsection{Multi-Teacher Distillation Across Interfaces}
\label{sec:cross-interface-consolidation}

We next combine T2I generation and I2I editing under UnifiedReward~\citep{wang2025unified,codegoat2025unifiedreward2} and CLIPScore with EditScore-4B~\citep{luo2026editscore}.
In Table~\ref{tab:cross-interface-reward-results}, each suite compares initialization, route teachers, midpoint merging~\citep{wortsman2022model,ilharco2022editing}, direct joint reinforcement learning, DanceOPD, and DiffusionOPD under equal-time and equal-data budgets.

\begin{table}[t]
\centering
\includegraphics[width=\textwidth]{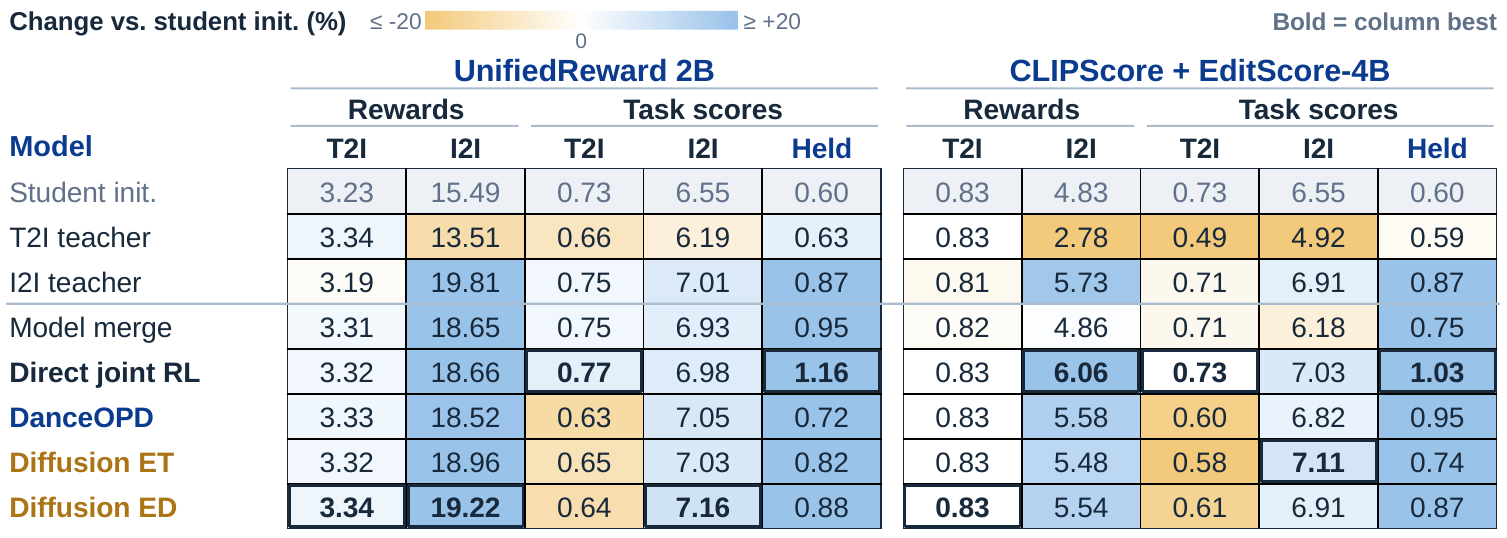}
\vspace{-0.42cm}
\caption{\textbf{Cross-interface distillation and joint reinforcement learning.} Direct joint reinforcement learning leads held-out preservation with online rewards, whereas equal-data DiffusionOPD leads recovery from frozen teachers.}
\label{tab:cross-interface-reward-results}
\end{table}

\noindent\textbf{Online joint reinforcement learning and distillation from frozen teachers complement each other.}
With online rewards, direct joint reinforcement learning preserves the broadest task profile.
Under UnifiedReward, it improves general T2I, general I2I, and held-out I2I by $5.5\%$, $6.6\%$, and $93.3\%$.
Similarly, under CLIPScore with EditScore-4B, it keeps general T2I unchanged while improving general I2I, EditScore, and held-out I2I by $7.3\%$, $25.5\%$, and $71.7\%$.
Conversely, without online rewards, equal-data DiffusionOPD leads combined T2I reward, I2I reward, and general I2I under UnifiedReward, exceeding direct joint reinforcement learning by $2.6\%$ on general I2I, and achieves the strongest frozen-teacher target reward combination across both suites.
Meanwhile, midpoint merging works under UnifiedReward where teachers share one reward family, yet degrades general T2I and I2I under heterogeneous teachers.
Notably, because standalone single-reward T2I teachers in Table~\ref{tab:cross-interface-reward-results} score below initialization on general T2I ($0.66$ and $0.49$ versus $0.73$), their T2I transfer reflects Rule 1, while equal-data DiffusionOPD still leads frozen-teacher editing recovery ($7.16$) and combined target reward transfer.
Unlike aligned merging~\citep{matena2022fisher,wu2026offpolicymerging}, distillation combines heterogeneous frozen checkpoints directly in Figure~\ref{fig:e5b-i2i-grid}.

\subsection{Asymmetric Prefix On-Policy Distillation With PrefixOPD}
\label{sec:target-strength}
\label{sec:clx-method}
\label{sec:clx-evaluation}

Existing pipelines apply symmetric queries at low noise in DanceOPD~\citep{zhou2026danceopd}, dense transition queries in DiffusionOPD~\citep{li2026diffusionopd}, phase-selective skipping within a shared regime in PhaseOPD~\citep{yan2026phaseopd}, or full-rollout extrapolation and block alignment in GOPD, DreOPD, and STEP-OPD~\citep{yang2026gopd,lin2026dreopd,wei2026stepopd}.
Section~\ref{sec:mechanism} shows that symmetric supervision mismatches generation-editing timing, runs redundant suffix steps, and under-utilizes early teacher updates.
PrefixOPD therefore pairs three components via Equations~\ref{eq:shared-extrapolation} and~\ref{eq:clx-loss}:
\begin{enumerate}
\item \textbf{Task-asymmetric query placement.} PrefixOPD keeps T2I queries at low-noise steps where solver intervals expand by $5.75\times$, and places I2I queries at step $1$ where edit direction forms.
\item \textbf{Prefix rollout truncation.} Because our causal ODE solver, suffix-independent query sets $\mathcal I_r$, and detached targets do not depend on steps $k>k_{\max}$, terminating each rollout after $k_{\max}$ exactly preserves the corresponding full-rollout stop-gradient update in Equation~\ref{eq:clx-loss} (holding $p_\theta$ and visited states fixed), cutting I2I student forwards by $81.8\%$ (from $10+1$ to $1+1$ rollout/gradient forwards, skipping $8$ suffix steps and the detached $k=1$ forward, i.e., $9$ of $11$).
\item \textbf{Initialization-referenced target extrapolation.} Using student initialization directly as the frozen editing reference in Equation~\ref{eq:shared-extrapolation}, PrefixOPD keeps T2I targets unchanged and applies $25\%$ extrapolation to amplify specialist updates on early I2I queries.
\end{enumerate}

We evaluate PrefixOPD on SD3.5-M with the T2I specialist and UnifiedReward-Edit (which satisfies Rules 1 and 2, exceeding initialization by $7.0\%$ on general I2I and $45.0\%$ on held-out I2I) across all $13$ routes and the three-route editing diagnostic suite (Appendices~\ref{app:benchmark-protocol}, \ref{app:clx-protocol}, and~\ref{app:target-strength-protocol}).

\begin{figure}[t]
\centering
\vspace{-0.2cm}
\includegraphics[width=\textwidth]{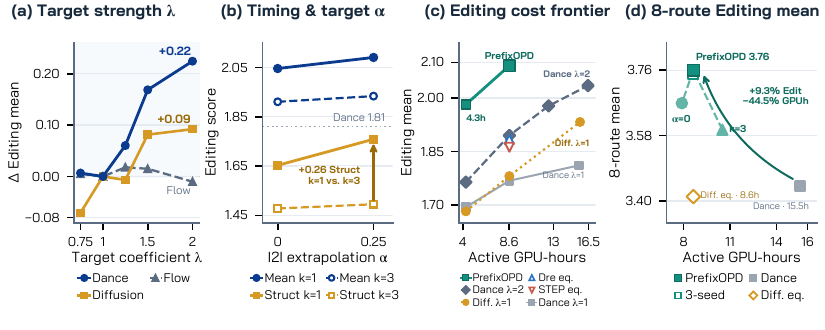}
\vspace{-0.65cm}
\caption{\textbf{Target extrapolation and PrefixOPD results.} (a) Extrapolation improves deterministic hosts while Flow-OPD stays flat, and (b) early queries drive most gains on (c) the cost frontier and (d) overall editing.}
\label{fig:target-strength}
\label{fig:clx-ablation}
\label{fig:clx-cost}
\end{figure}

\begin{table}[t]
\centering
\vspace{-0.12cm}
\includegraphics[width=\textwidth]{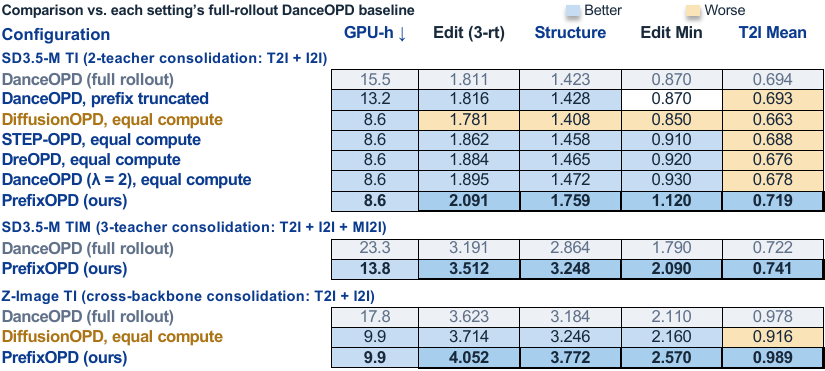}
\vspace{-0.26cm}
\caption{\textbf{Three-route editing diagnostic suite and generation benchmarks.} PrefixOPD leads Edit (3-rt), structure, worst-case, and T2I scores at lower cost across mixtures and backbones, surpassing equal-data full-rollout extrapolation in Table~\ref{tab:lambda-complete} ($2.035$ DanceOPD $\lambda=2$, $16.4$ GPU-h; $2.025$ DiffusionOPD $\lambda=2$) at $2.091$ ($8.6$ GPU-h).}
\label{tab:clx-primary}
\vspace{-0.15cm}
\end{table}

\noindent\textbf{Target extrapolation benefits deterministic hosts but not clipped stochastic distillation.}
In Figure~\ref{fig:target-strength}(a) and Table~\ref{tab:lambda-complete}, $\lambda=2$ lifts Editing Mean by $12.4\%$ on DanceOPD and $4.8\%$ on DiffusionOPD while $\lambda<1$ lowers Editing Min.
Flow-OPD stays within $1.1\%$ across $\lambda$ as shifted targets conflict with ratio clipping and initialization anchoring, and 8-bit quantized references match float16 within $0.7\%$.

\noindent\textbf{Early editing queries drive most gains and generalize across settings.}
In Figures~\ref{fig:target-strength}(b) and~\ref{fig:target-strength}(c) and Table~\ref{tab:clx-all-ablations}, truncating DanceOPD after its last query reduces active time by $14.8\%$ within $0.3\%$ score difference.
Relative to this truncated control, PrefixOPD reduces active time by a further $34.8\%$ while improving Editing Mean by $15.1\%$ and T2I Mean by $3.8\%$, with three-seed means confirming $+15.5\%$ Editing Mean and $+2.6\%$ T2I Mean gains over DanceOPD.
Moreover, querying I2I at step $1$ without extrapolation alone reduces active time by $49.3\%$ over DanceOPD and improves Editing Mean by $13.0\%$, Editing Structure by $16.2\%$, and Editing Min by $20.7\%$.
This single timing shift accounts for $83.9\%$ of PrefixOPD's editing gain, whereas later or denser early queries reduce Editing Structure by up to $15.0\%$.
Conversely, moving T2I queries to step $1$ under the shifted schedule reduces T2I Mean by $5.6\%$, while uniform T2I steps reverse preference toward step $1$ by $1.9\%$ with editing gains intact.
Swapping teachers in Tables~\ref{tab:teacher-strength} and~\ref{tab:clx-all-ablations} further confirms Rules 1 and 2, as local supervised fine-tuning exceeds local reinforcement learning by $5.6\%$, UnifiedReward-Edit exceeds local supervised fine-tuning by $7.5\%$, and EditScore-4B improves Editing Mean by $14.0\%$ under PrefixOPD.
Additionally, PrefixOPD retains $14.8\%$ and $15.5\%$ Editing Mean leads across unguided and guided inference and reduces active time by $40.8\%$ with a $10.1\%$ Editing Mean gain on SD3.5-M TIM.
On Z-Image TI, PrefixOPD similarly reduces active time by $44.4\%$ while outperforming DanceOPD by $11.8\%$ and DiffusionOPD at equal compute by $9.1\%$.

\noindent\textbf{PrefixOPD establishes a superior quality-efficiency frontier across benchmarks and evaluators.}
In Table~\ref{tab:clx-primary}, PrefixOPD reduces active training time by $44.5\%$ over DanceOPD while improving Editing Mean by $15.5\%$, Editing Structure by $23.6\%$, and Editing Min by $28.7\%$.
At half budget in Figure~\ref{fig:target-strength}(c), PrefixOPD already exceeds full-rollout DanceOPD by $9.4\%$ in roughly a quarter of the time.
At equal compute, PrefixOPD outperforms extrapolated DanceOPD, DreOPD, STEP-OPD, and DiffusionOPD by up to $17.4\%$ on Editing Mean and $8.4\%$ on T2I Mean; even against equal-data full-rollout $\lambda=2$ extrapolation in Table~\ref{tab:lambda-complete} ($2.035$ DanceOPD at $16.4$ GPU-hours; $2.025$ DiffusionOPD), PrefixOPD ($2.091$, $8.6$ GPU-hours) leads Editing Mean ($+2.8\%$ to $+3.3\%$) and Editing Min ($+8.7\%$ to $+14.3\%$) at nearly half the cost.
Across all $13$ routes in Figure~\ref{fig:target-strength}(d) and Appendix~\ref{app:clx-full13-sub}, PrefixOPD improves T2I Mean by $3.7\%$ ($3.6\%$ at three-decimal table precision; $+2.6\%$ across three seeds in Table~\ref{tab:clx-all-ablations}), trained I2I Mean by $16.2\%$, and overall eight-route Editing Mean by $9.3\%$.
These gains yield a $66.5\%$ win rate in Figure~\ref{fig:prefixopd-qualitative}, plus $55.6\%$ higher T2I text over initialization and a $2.2\%$ TIM composition lead.
Appendix~\ref{app:extension-judging} confirms $85.9\%$ pairwise winner agreement with InternVL2.5-8B across all $32$ endpoints ($84.1\%$ on the six-endpoint audit).

\section{Conclusion}
Diffusion on-policy distillation transfers specialist strengths selectively rather than copying teachers uniformly.
Across budgets and trajectories, query count drives $83.0\%$ of equal-data gains, early high-noise steps capture $80.2\%$ of the final edit direction, and $5.75\times$ wider late solver steps favor low-noise generation queries.
Moreover, narrow single-route reinforcement learning teachers without cross-route regularization degrade student held-out editing despite higher target scores, whereas joint reinforcement learning and frozen-teacher distillation serve complementary reward settings.
Grounded in these mechanisms, PrefixOPD reduces active training time by $44.5\%$ while improving text-to-image generation by $3.7\%$, overall eight-route editing by $9.3\%$, and the three-route editing diagnostic suite by $15.5\%$, exceeding equal-compute baselines by up to $17.4\%$.
We hope our mechanistic findings and PrefixOPD inspire efficient multi-teacher post-training across visual and multimodal generative models.

\clearpage

\appendix
\raggedbottom

\begingroup
\setcounter{tocdepth}{2}
\section*{Appendix Contents}
\makeatletter
\@starttoc{atoc}
\makeatother
\endgroup
\vspace{0.35cm}

\begingroup
\let\appendixaddcontentsline\addcontentsline
\renewcommand{\addcontentsline}[3]{\ifstrequal{#1}{toc}{\appendixaddcontentsline{atoc}{#2}{#3}}{\appendixaddcontentsline{#1}{#2}{#3}}}
\section{Related Work}
\label{sec:related-work}

\noindent\textbf{On-policy distillation in diffusion models.}
On-policy distillation supervises a student at states visited along its own denoising trajectories, whereas earlier diffusion distillation targeted faster sampling via trajectory imitation or distribution matching~\citep{luo2023diffinstruct,garrepalli2024ddil}.
In multi-teacher distillation, Flow-OPD combines routed supervision with manifold regularization, DiffusionOPD matches transition means, and DanceOPD matches sparse velocities, alongside multi-task LoRA and 3D distillation~\citep{wu2026collectionlora,ning2026flow3dopd}.
Other studies match continuous trajectories in FlowCTS~\citep{ye2026flowcts}, extrapolate targets in GOPD and DreOPD, or align blockwise feature changes in STEP-OPD.
Related methods select query phases and skip rollout steps in PhaseOPD~\citep{yan2026phaseopd}, or distill trajectories and residual proxies for few-step generation in $\pi$-Flow, SMD, and StepX-OPD~\citep{chen2026piflow,zhang2026smd,wang2026stepxopd}.
Complementary rollout methods match score or moment distributions~\citep{yin2024dmd,yin2024dmd2,zhou2024sid,zhou2025imm} or enforce self-consistency across video and flow models~\citep{huang2025selfforcing,gu2026anyflow,qin2026soar}.
Self-distillation variants use privileged conditioning or reward updates~\citep{jiang2026d,wang2026dopcd,zhang2026selfopd,jiang2026dmdr}, while hybrid methods couple distribution matching with reward gradients~\citep{chen2026flashdmd,dong2026gdmd,jacq2026rmmd}.
Further analyses study guidance branch errors~\citep{li2026cfgopd,zhang2026gfdopd} or bridge heterogeneous latent spaces~\citep{fu2026poly,fu2026anyopd}.
Meanwhile, diffusion reinforcement learning spans policy optimization~\citep{fan2023dpok,liu2025flowgrpo,xue2025dancegrpo,ye2025ddrl}, differentiable or temporal credit assignment~\citep{clark2024draft,he2026tempflow,li2026mixgrpo}, and reward regularization~\citep{hong2026rewardhacking,zhu2026crd,golan2026paretoslider}.
Building on student-state velocity matching and target extrapolation under fixed inference schedules, we establish how query count, asymmetric timing, and teacher preservation govern multi-teacher transfer in PrefixOPD.

\noindent\textbf{Knowledge transfer across teachers and students.}
How teacher behaviors transfer has been studied from classic model compression to modern generative distillation.
Self-distillation demonstrates that equal-capacity students can surpass their teachers~\citep{furlanello2018ban}, while theoretical and empirical analyses show that distillation absorbs complementary multi-view ensemble features~\citep{allenzhu2023understanding}, that student generalization often diverges from teacher fidelity~\citep{stanton2021does}, and that training consistency and schedule budget strongly govern transfer~\citep{beyer2022knowledge}.
Because stronger teachers can widen optimization or capacity gaps~\citep{cho2019efficacy,busbridge2025distillation,li2025small}, prior work introduces teacher assistants~\citep{mirzadeh2020improved} and temporally adaptive target interpolation~\citep{shing2025taid}.
In autoregressive language models, reverse-KL and adaptive on-policy objectives reduce exposure bias~\citep{gu2024minillm,ko2024distillm}, while recent analyses examine teacher hacking~\citep{tiapkin2025teacherhacking}, state coverage and data-versus-algorithm dynamics~\citep{li2026rethinking}, and multi-teacher capability consolidation~\citep{xie2026multiteacher}.
In multi-teacher diffusion distillation, FocusOPD~\citep{lin2026focusopd} finds that raw feature discrepancy peaks at early high-noise steps and deep blocks.
Under output-level trajectory distillation, we separate raw feature discrepancy from localized CKA shifts over initialization, solver-amplified low-noise updates, and cross-route teacher preservation.

\section{Training And Evaluation Protocols}
\label{app:implementation}

\subsection{Supervised Fine-Tuning}
\label{app:teacher-sft}

\noindent\textbf{Backbones and optimization.}
We train supervised fine-tuning specialists from SD3.5-M~\citep{stability2024sd35medium,esser2024scaling} and Z-Image~\citep{cai2025z} at $512\times512$ resolution on $8$ NVIDIA H100-80GB GPUs using batch size $32$, rank-$32$ LoRA, frozen encoders, and AdamW with learning rate $10^{-4}$, weight decay $10^{-4}$, gradient clip $1.0$, and $100$ warm-up steps.
Z-Image uses uniform flow timesteps, sequence length $512$, and attention LoRA, evaluated at $50$ steps with classifier-free guidance (CFG) $4.0$.
SD3.5-M uses logit-normal timesteps, CLIP-L, CLIP-G, and T5 encoders at length $256$, joint and context attention LoRA, and learned edit embeddings, evaluated at $40$ steps with guidance $4.5$.

\noindent\textbf{T2I, I2I, and MI2I specialists.}
The $5$ T2I routes draw $100{,}000$ deduplicated examples each from Fine-T2I~\citep{ma2026fine} and TextGround4M~\citep{mao2026textground4m}, trained for $1$ epoch of $3{,}125$ updates.
The I2I pool curates $1.95\mathrm{M}$ triplets from GPT-Image-Edit, Pico-Banana, and Nano-consistent-150k~\citep{wang2025gpt,qian2026pico,ye2025nanoconsistent} together with ShareGPT-4o-Image, ScaleEdit-12M, and UnicEdit-10M~\citep{chen2025sharegpt,chen2026scaleedit,ye2026unicedit}, training a general parent before warm-starting local, global, and text specialists with loss only on the target latent.
The MI2I pool draws $510\mathrm{K}$ examples from MACRO, Echo-4o, and MICo-150K~\citep{chen2026macro,ye2025echo,wei2026mico} plus OmniRef~\citep{huang2026scaling}, training spatial, identity, and composition specialists with up to $10$ clean role-labeled references under a $512^2$ pixel budget.

\subsection{Reinforcement Learning}
\label{app:teacher-rl}

Each single-route reinforcement specialist starts from its supervised counterpart and optimizes one per-route reward from frozen InternVL3.5-38B~\citep{wang2025internvl3} (without cross-route regularization) using on-policy diffusion policy updates~\citep{zhou2026policy}; the same vision-language model~\citep{chow2025physbench} partitions unlabeled editing pairs during curation.
Updates draw $100$ conditions and $N_{\mathrm{cand}}=6$ candidates over $10$ steps at CFG $1.0$ via AdamW (learning rate $3\times10^{-4}$), maximizing $\widehat R_{\mathrm{exp}}=\sum_{c=1}^{5}c\,\pi_c$ with $\pi_c=\frac{\exp(l_c)}{\sum_{h=1}^{5}\exp(l_h)}$ for score-token logits $l_c$, selecting peak validation reward on a disjoint split.

\subsection{Common On-Policy Distillation Design And Diagnostics}
\label{app:opd-common-design}
\label{app:absorption-diagnostics}

\noindent\textbf{Mixtures and shared controls.}
TI combines general T2I and general I2I over $3{,}000$ conditions per route for $188$ updates starting from $S_0$ initialized at the global I2I reinforcement endpoint, while TIM adds MI2I composition for $282$ updates from an equal-weight merge of the global I2I and composition endpoints.
Initializing $S_0$ from post-trained specialist checkpoints models practical post-training consolidation where high-reward endpoints exhibit cross-route forgetting on unoptimized text and held-out routes, evaluating multi-teacher recovery without retraining from base backbones.
DanceOPD, DiffusionOPD, and Flow-OPD share data, $S_0$, rank-$32$ LoRA with $\alpha_{\mathrm{LoRA}}=64$, optimizer, and $10$-step unguided rollouts.

\noindent\textbf{Behavioral, CKA, and DINOv2 diagnostics.}
For model $f$ on held-out route $e$ with $n_e=100$ examples and $J_e$ rubric scores $q_{e,i,j}(f)$, the per-image minimum score is $M_e(f)\coloneqq \frac{1}{n_e}\sum_{i=1}^{n_e}\min_{1\leq j\leq J_e}q_{e,i,j}(f)$.
Linear CKA evaluates $32$ general and held-out I2I conditions at sampler steps $1$, $5$, and $9$ across $6$ uniform transformer depths.
Using frozen DINOv2 ViT-B/14 with registers at $518$ pixels, we compute patch shifts $\Delta\phi_k\coloneqq\phi(\hat I_k)-\phi(I^{\mathrm{src}})$ and mask-pooled shifts $\Delta\bar\phi_k\coloneqq\bar\phi(\hat I_k)-\bar\phi(I^{\mathrm{src}})$ relative to $\Delta\bar\phi^{\mathrm{gt}}\coloneqq\bar\phi(I^{\mathrm{gt}})-\bar\phi(I^{\mathrm{src}})$ on top-$25\%$ divergent mask $M^{\mathrm{gt}}$ for $\bar\phi(\cdot)\coloneqq\frac{1}{|M^{\mathrm{gt}}|}\sum_{t\in M^{\mathrm{gt}}}\phi_t(\cdot)$ and ground-truth image $I^{\mathrm{gt}}$.

\subsection{Teacher Performance And Reward Protocols}
\label{app:teacher-performance}
\label{app:reward-evaluation}

Table~\ref{tab:teacher-performance} reports all $44$ paired supervised and reinforcement specialists across the $13$ routes.
Within T2I, PickScore, CLIPScore, and HPSv2.1 teachers are compared with single-teacher distillation, joint distillation, and normalized direct reinforcement learning.
Cross-interface suites pair UnifiedReward-2.0-2B with UnifiedReward-Edit-2B and CLIPScore with EditScore-4B on disjoint test splits.

\begin{table*}[tp]
\centering
\caption{\textbf{Specialist performance across benchmark routes.} Paired supervised and reinforcement teachers are evaluated across generation, editing, and multi-image routes.}
\label{tab:teacher-performance}
\includegraphics[width=\textwidth]{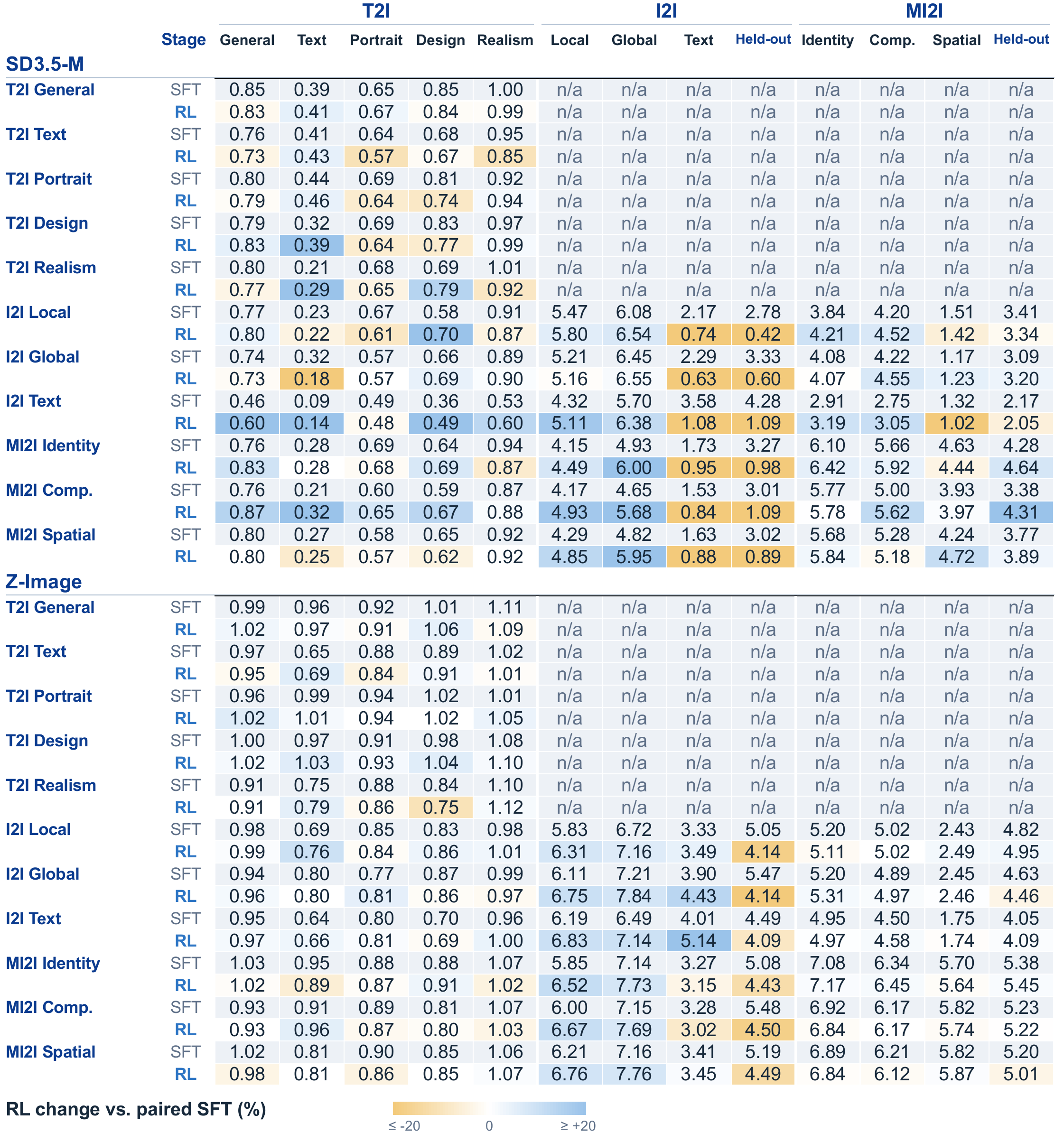}
\end{table*}

\section{Evaluation}
\label{app:teacher-evaluation}

\subsection{Benchmark And Shared Evaluation Protocol}
\label{app:benchmark-protocol}
Each of the $5$ T2I, $4$ I2I, and $4$ MI2I routes evaluates $100$ fixed test conditions disjoint from training and validation.
T2I routes use Qwen-Image-Bench~\citep{li2026qwen} scored by Q-Judger at temperature $0$.
Complementing human-preference edit reward data~\citep{wu2025editreward,bai2024humanedit,yu2025anyedit}, local and global I2I use GEdit-Bench~\citep{liu2025step1x}, text I2I uses WeEdit~\citep{zhang2026weedit}, identity and spatial MI2I use MACRO, composition uses MultiBanana~\citep{oshima2026multibanana}, and held-out I2I and MI2I each use $100$ unseen multi-family conditions scored by GPT-5.5 at temperature $0$ on $[0,10]$ with condition mean $q_{e,i}=\frac{1}{J_e}\sum_{j=1}^{J_e} q_{e,i,j}$.
On the three representative editing routes $\mathcal P=\{\text{I2I-Text},\text{MI2I-Spatial},\text{MI2I-Held-out}\}$ with structural criteria $\mathcal S_e\subset\{1,\ldots,J_e\}$, we report
\begin{align*}
\mathrm{Editing\ Mean}&=\frac{1}{|\mathcal P|}\sum_{e\in\mathcal P}\frac{1}{n_e}\sum_{i=1}^{n_e} q_{e,i}, &
\mathrm{Editing\ Min}&=\frac{1}{|\mathcal P|}\sum_{e\in\mathcal P}\frac{1}{n_e}\sum_{i=1}^{n_e}\min_{1\leq j\leq J_e} q_{e,i,j},\\
\mathrm{Editing\ Structure}&=\frac{1}{|\mathcal P|}\sum_{e\in\mathcal P}\frac{1}{n_e}\sum_{i=1}^{n_e}\frac{1}{|\mathcal S_e|}\sum_{j\in\mathcal S_e} q_{e,i,j}. & &
\end{align*}

\section{Detailed Results}
\label{app:complete-scores}
\label{app:complete-heldout}
\label{app:complete-temporal}
\label{app:complete-teacher-source}
\label{app:complete-reward}

Across the $36$ held-out comparisons under the per-image minimum score $M_e(f)$, students exceed their routed teachers in $72.2\%$ of cells, with positive rates of $91.7\%$ for Flow-OPD, $66.7\%$ for DanceOPD, and $58.3\%$ for DiffusionOPD, reversing the individual-criterion ordering of DanceOPD and DiffusionOPD.

\begin{table}[!htbp]
\centering
\caption{\textbf{Temporal supervision across all $13$ routes.} Fixed query schedules and solver-step controls.}
\label{tab:temporal-scheduling}
\includegraphics[width=\textwidth]{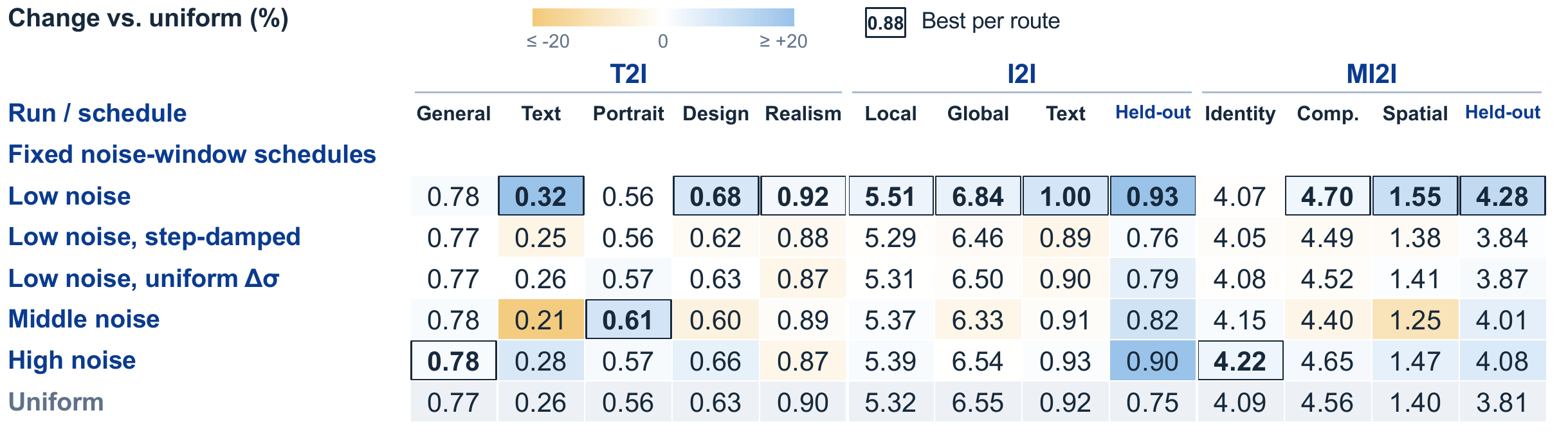}
\end{table}

\begin{table}[!htbp]
\centering
\caption{\textbf{Teacher-source transfer across routes and backbones.} Initialization, paired teachers, and DanceOPD students.}
\label{tab:teacher-strength}
\includegraphics[width=\textwidth]{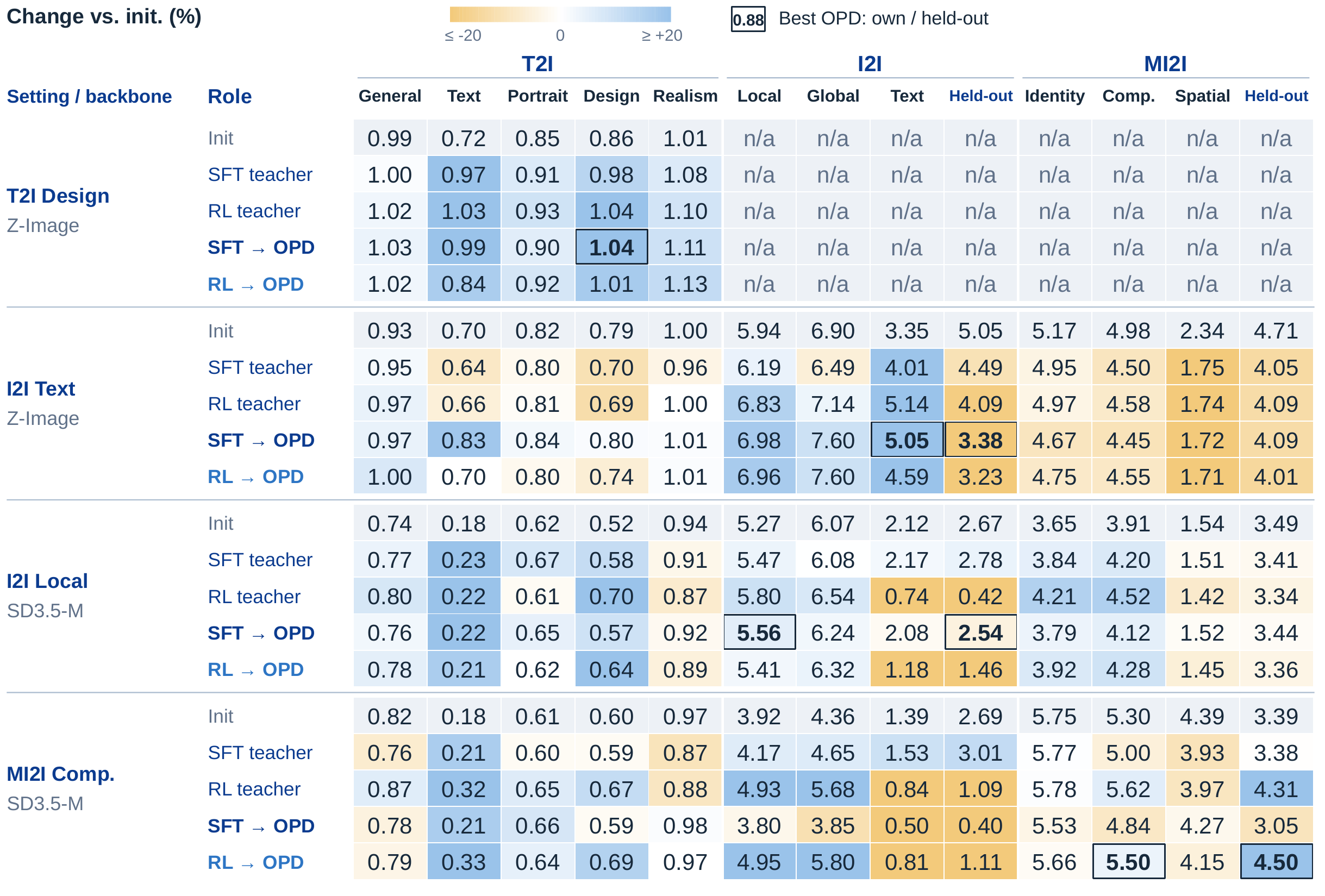}
\end{table}

\section{More Results}
\label{app:supplementary}

\subsection{Budget Transitions And Representation Examples}
\label{app:representation-cases}
Figure~\ref{fig:budget-transitions} reports all budget winner transitions and the query-density ablation, and Figure~\ref{fig:consolidation-qualitative} shows matched outputs across initialization, specialists, and equal-data students.
Figure~\ref{fig:cka-depth-ranges} summarizes CKA profiles across sampler states and depths, while static DINOv2 reference-patch similarity remains dominated by unedited background tokens across distilled students.

\begin{figure}[t]
\centering
\includegraphics[width=\textwidth]{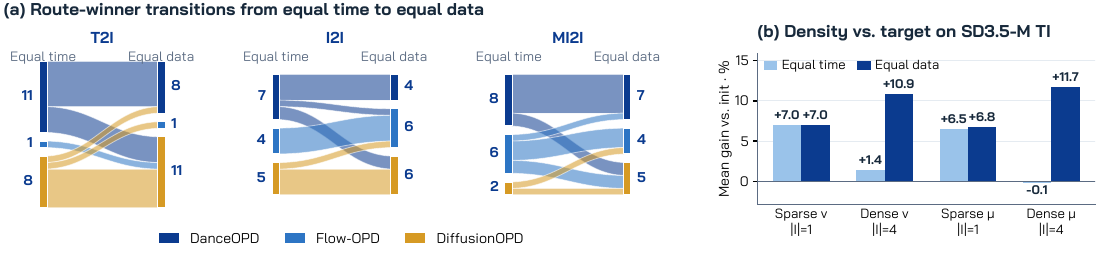}
\vspace{-0.48cm}
\caption{\textbf{Budget winner transitions and query-density ablation.} (a) Winner shifts from equal time to equal data, and (b) isolation of query count per trajectory.}
\label{fig:budget-transitions}
\end{figure}

\begin{figure}[!htbp]
\centering
\includegraphics[width=\textwidth,height=0.98\textheight,keepaspectratio]{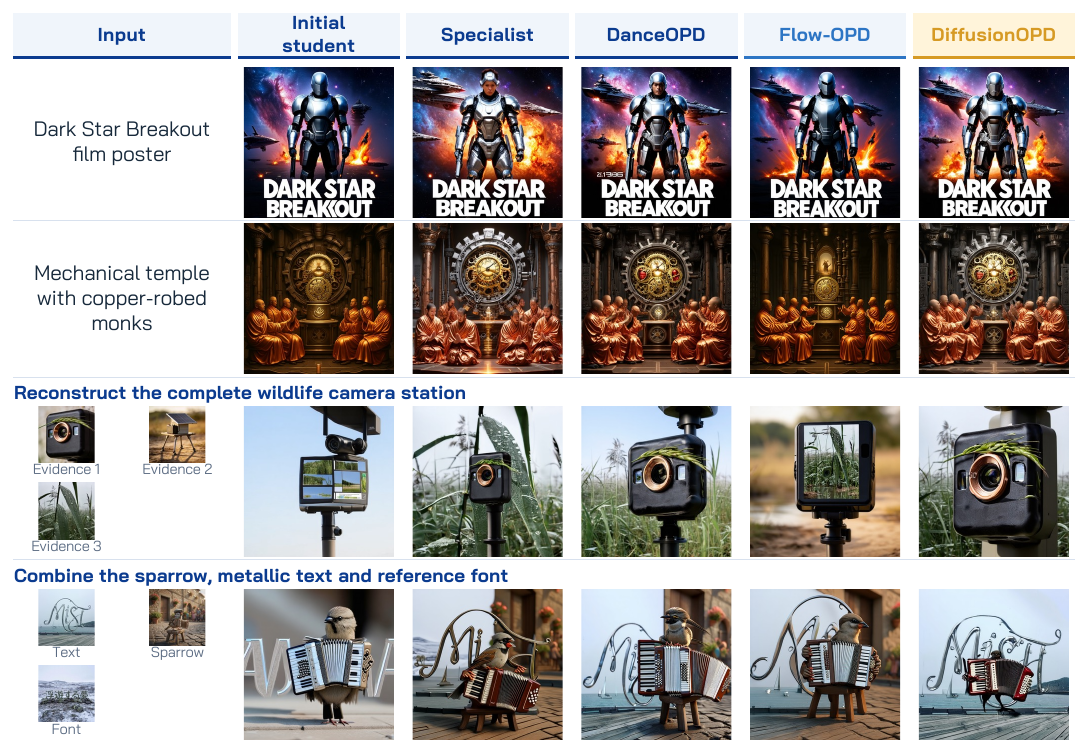}
\vspace{-0.50cm}
\caption{\textbf{Qualitative comparison of multi-teacher distillation.} Matched SD3.5-M TIM cases across initialization, specialists, and equal-data students.}
\label{fig:consolidation-qualitative}
\end{figure}

\begin{figure}[t]
\centering
\includegraphics[width=\textwidth]{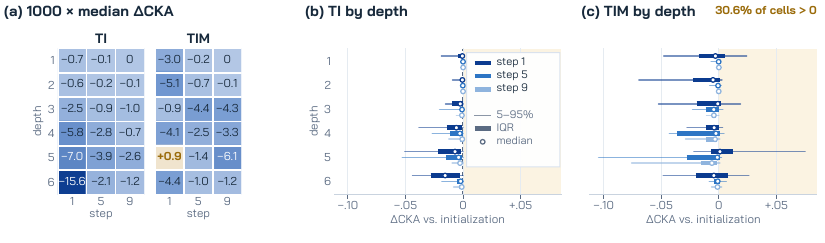}
\vspace{-0.68cm}
\caption{\textbf{Representation resemblance across states and depth.} (a) Median CKA profiles on (b) TI and (c) TIM show localized positive alignment in deeper decoding blocks (notably at step $1$).}
\label{fig:cka-depth-ranges}
\end{figure}

\subsection{Trajectory Interventions And Temporal Training}
\label{app:temporal-training}
\label{app:trajectory-interventions}

To complement the $48$ stratified cases in Section~\ref{sec:frozen-trajectory} and Figure~\ref{fig:trajectory-interventions}, we evaluate $32$ additional I2I conditions where $20\%$ interventions yield $2.18\times$ and $3.45\times$ the gains of $50\%$ and $80\%$ interventions.
Early single steps exceed late steps by $66.7\%$ on local edits and $2.75\times$ on text edits and close over $76\%$ of large-gap spatial failures, while edited-patch DINOv2 alignment exceeds T2I by $2.81\times$ initially and $2.30\times$ at step $1$.

Low-, middle-, and high-noise schedules in Table~\ref{tab:temporal-scheduling} sample progress from Beta distributions $\mathrm{Beta}(5,2)$, $\mathrm{Beta}(5,5)$, and $\mathrm{Beta}(2,5)$.
In Figures~\ref{fig:e4-solver-probe}(c) and~\ref{fig:e4-solver-probe}(f), shifted schedules keep low-to-high update ratios at $1.75\times$ on midpoint, $1.57\times$ on Euler, $1.43\times$ on Heun-2, and $1.48\times$ on the rollout solver; damped weighting reduces low-noise AdamW cosine by $41.9\%$ and error reduction by $33.6\%$, whereas squared step weighting on fixed states raises late-step AdamW cosine by $12.8\times$ and error reduction by $3.89\times$.

\subsection{Teacher-Source And Reward Outputs}
\label{app:teacher-source-audit}
\label{app:reward-visuals}
Figure~\ref{fig:teacher-source-expanded} shows matched multi-image composition outputs across supervised and reinforcement teachers and their DanceOPD students, complementing Table~\ref{tab:teacher-strength} where early relative velocity disagreement exceeds late disagreement by up to $8.66\times$ across all $44$ specialists.
Figures~\ref{fig:e5b-t2i-grid} and~\ref{fig:e5b-i2i-grid} present matched T2I and I2I outputs across initialization, reward specialists, parameter merging, direct joint reinforcement learning, DanceOPD, and DiffusionOPD under cross-interface reward suites.

\begin{figure}[!htbp]
\centering
\includegraphics[width=\textwidth,height=0.98\textheight,keepaspectratio]{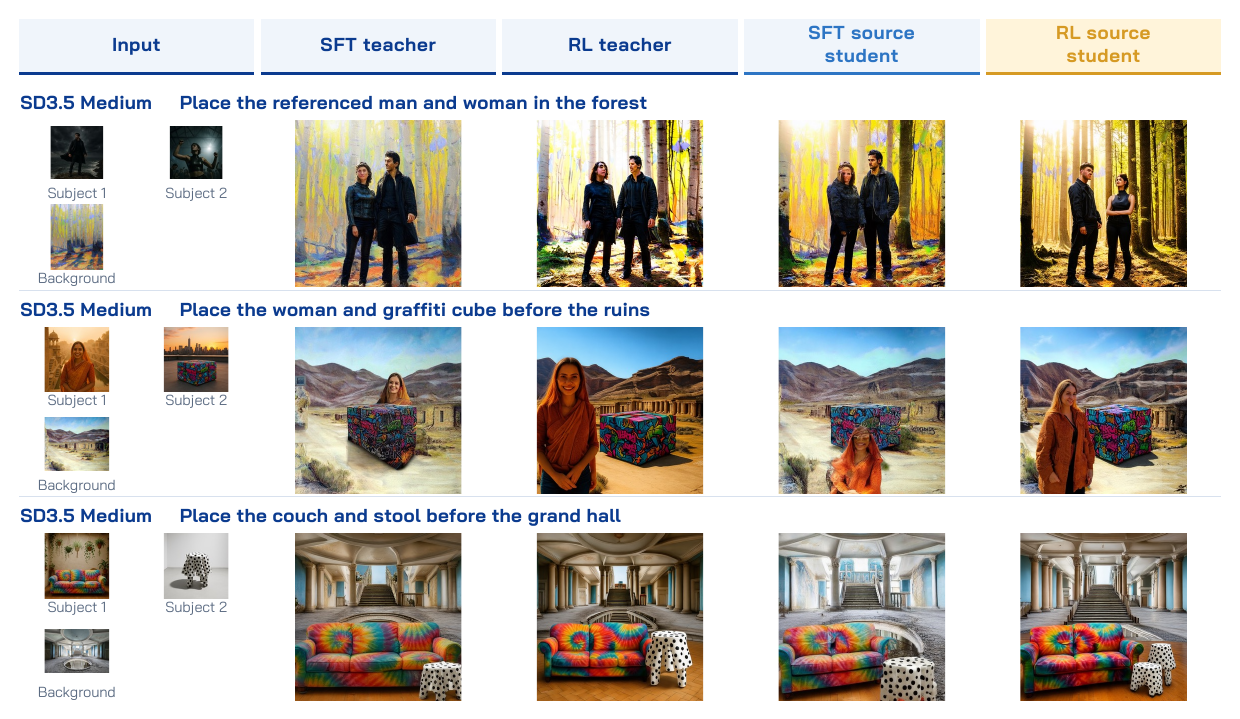}
\vspace{-0.52cm}
\caption{\textbf{Qualitative effects of teacher source.} Matched multi-image composition outputs from supervised and reinforcement teachers and their DanceOPD students.}
\label{fig:teacher-source-expanded}
\end{figure}

\begin{figure}[!htbp]
\centering
\includegraphics[width=\textwidth,height=0.98\textheight,keepaspectratio]{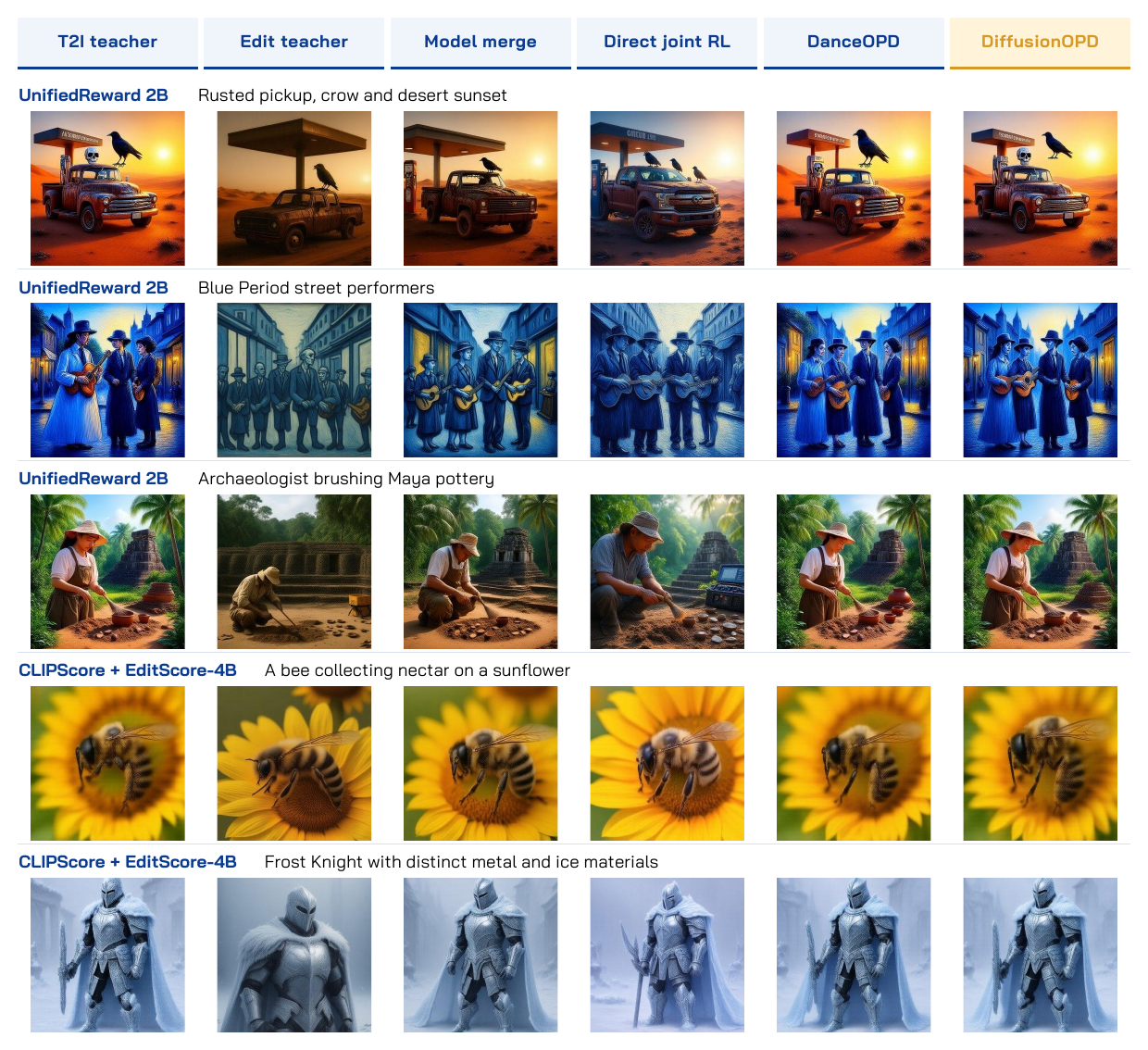}
\vspace{-0.55cm}
\caption{\textbf{Text-to-image outputs under cross-interface reward suites.} Matched prompts across initialization, reward teachers, parameter merging, joint reinforcement learning, DanceOPD, and DiffusionOPD.}
\label{fig:e5b-t2i-grid}
\end{figure}

\begin{figure}[!htbp]
\centering
\includegraphics[width=\textwidth,height=0.98\textheight,keepaspectratio]{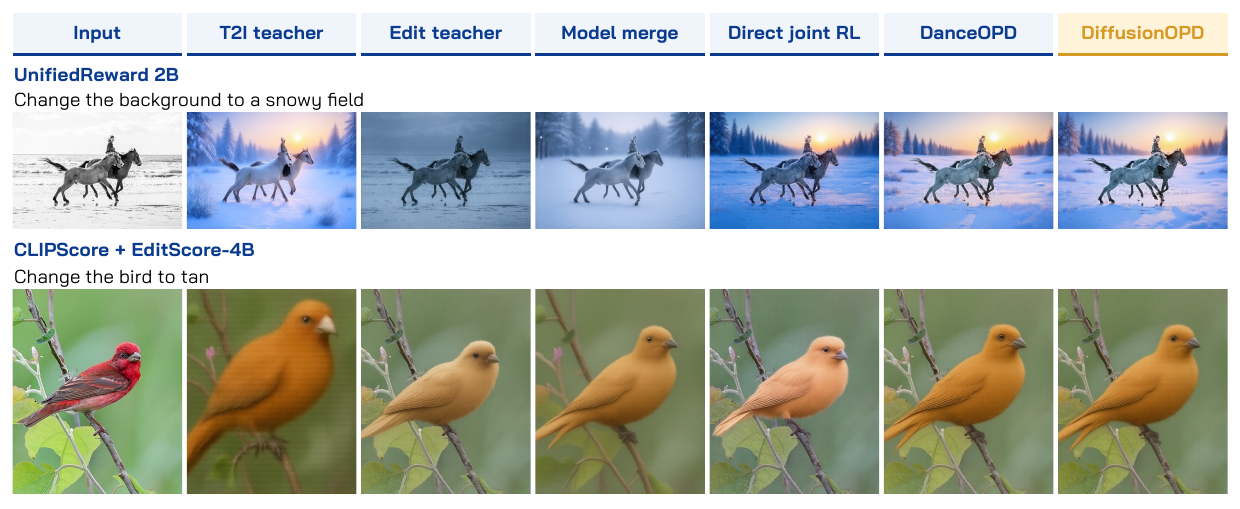}
\vspace{-0.50cm}
\caption{\textbf{Image-editing outputs under cross-interface reward suites.} Matched editing cases across reward teachers, parameter merging, joint reinforcement learning, DanceOPD, and DiffusionOPD.}
\label{fig:e5b-i2i-grid}
\end{figure}

\clearpage
\section{PrefixOPD Protocol And Complete Ablations}
\label{app:clx-protocol}

\subsection{Training Configuration And Inference Schedule Transfer}
\label{app:clx-setting}

On SD3.5-M from the editing initialization, PrefixOPD combines the T2I specialist and UnifiedReward-Edit over $3{,}000$ T2I and $3{,}000$ I2I conditions for $188$ updates using rank-$32$ LoRA, with student initialization serving as the frozen editing reference in Equation~\ref{eq:shared-extrapolation}.
Training runs $10$-step ODE rollouts at guidance $1.0$ and standard evaluation runs $40$ steps at guidance $4.5$, preserving late-to-early step ratios above $5.48\times$ and maintaining Editing Mean leads up to $15.5\%$ across step counts in Table~\ref{tab:inference-schedule-transfer}.
Synchronized active GPU-hours across $8$ NVIDIA H100-80GB GPUs average $8.63$ GPU-hours for PrefixOPD on SD3.5-M TI, reducing active time by $44.5\%$ over DanceOPD and $47.5\%$ over DanceOPD ($\lambda=2$), alongside $40.8\%$ reduction on SD3.5-M TIM and $44.4\%$ reduction on Z-Image TI.

\begin{table}[t]
\centering
\includegraphics[width=\textwidth]{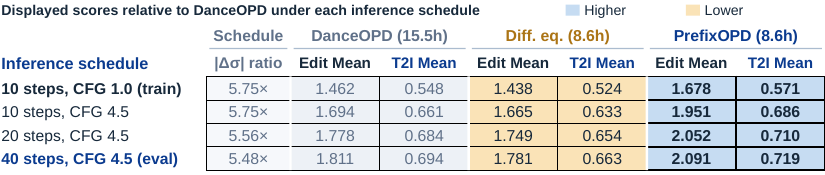}
\caption{\textbf{Transfer across inference step counts and guidance scales.} PrefixOPD preserves its generation and editing lead across unguided and multi-step guided sampling.}
\label{tab:inference-schedule-transfer}
\end{table}

\subsection{Component Ablations And Route-Level Gains}
\label{app:clx-full13-sub}

Table~\ref{tab:clx-all-ablations} reports component ablations where three-seed Editing Mean standard deviations stay below $0.3\%$ of the mean on the fixed $100$-condition-per-route test splits, and where initialization-referenced extrapolation ($\alpha=0.25$) further stabilizes shared LoRA updates against cross-route T2I drift relative to unextrapolated step-$1$ editing ($\alpha=0$).
All equal-compute baselines share the same $S_0$, data stream, rank-$32$ LoRA, and $8.63$ GPU-hour budget (with per-channel 8-bit quantized-reference target extrapolation for DreOPD and six-block feature alignment for STEP-OPD); at equal $8.63$ GPU-hours, PrefixOPD outperforms equal-compute DanceOPD ($\lambda=2$) by $10.3\%$, quantized DreOPD by $11.0\%$, six-block STEP-OPD by $12.3\%$, and DiffusionOPD by $17.4\%$ on Editing Mean while leading T2I Mean by up to $8.4\%$, and still surpasses equal-data $\lambda=2$ extrapolation in Table~\ref{tab:lambda-complete} ($2.035$ for DanceOPD at $16.4$ GPU-hours and $2.025$ for DiffusionOPD).
UnifiedReward-Edit improves ground-truth DINOv2 alignment by $14.8\%$ and background preservation by $17.8\%$ over local reinforcement learning, while EditScore-4B improves Editing Mean by $14.0\%$ and Editing Structure by $15.9\%$.

\begin{table}[!htbp]
\centering
\includegraphics[width=0.96\textwidth]{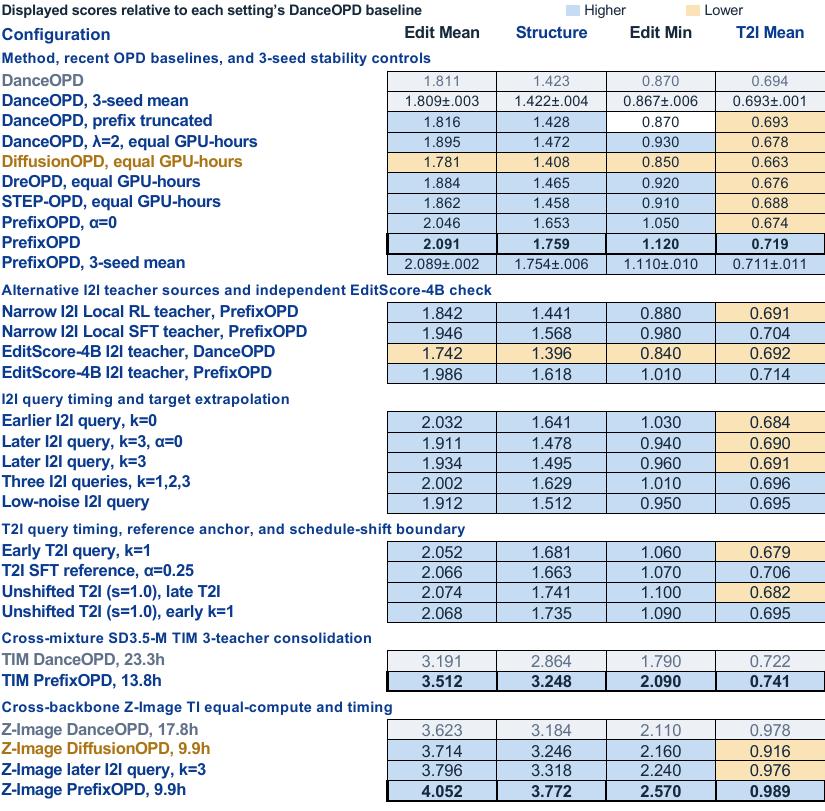}
\vspace{-0.12cm}
\caption{\textbf{Component ablations of PrefixOPD.} Controlled comparisons across truncations, equal-compute baselines, teachers, schedules, mixtures, and backbones.}
\label{tab:clx-all-ablations}
\end{table}

Across $10{,}000$ paired bootstrap draws over all $13$ routes, PrefixOPD reaches a $67.7\%$ non-tie win rate on the three-route editing diagnostic suite and $66.5\%$ over all $800$ editing conditions, improving held-out I2I by $44.3\%$, I2I text by $34.4\%$, MI2I spatial by $18.8\%$, local I2I by $17.3\%$, global I2I by $9.6\%$, and held-out MI2I by $8.4\%$ over DanceOPD in Figure~\ref{fig:prefixopd-qualitative}.
Versus equal-compute DiffusionOPD, PrefixOPD lifts T2I Mean by $8.4\%$ and overall eight-route Editing Mean by $10.2\%$, winning all $5$ T2I and $7$ of $8$ editing routes.

\begin{figure}[!htbp]
\centering
\includegraphics[width=\textwidth]{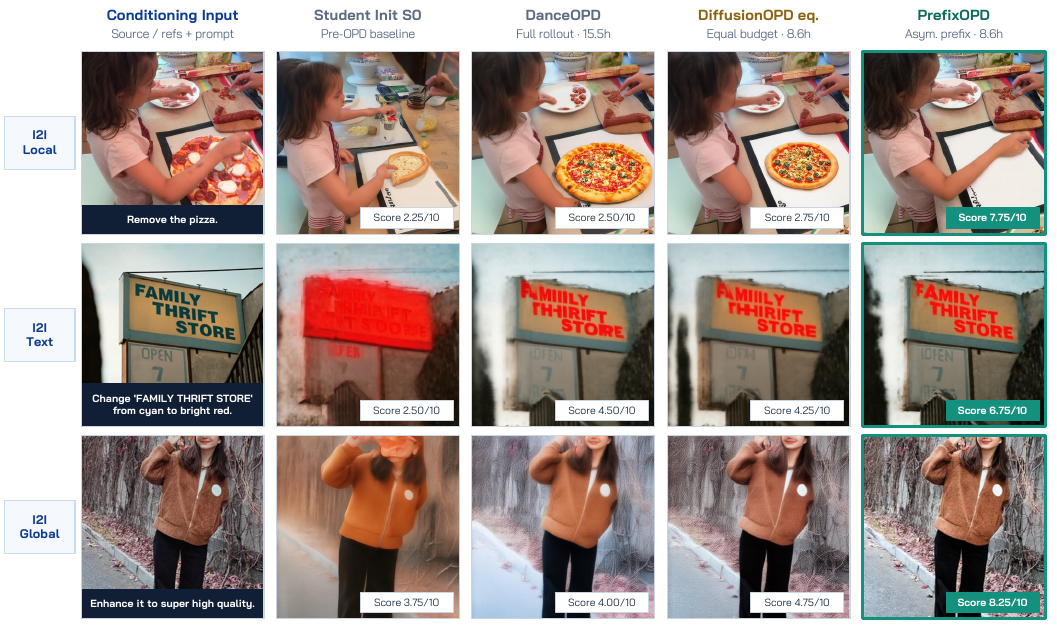}
\vspace{-0.38cm}
\caption{\textbf{Qualitative comparison of PrefixOPD and distillation baselines.} Matched examples across local object removal, text editing, and global restoration (showing $[0,10]$ editing scores).}
\label{fig:prefixopd-qualitative}
\end{figure}

\section{Cross-Method Target Strength And Evaluator Audit}
\label{app:target-strength-protocol}
\label{app:extension-judging}

Table~\ref{tab:lambda-complete} reports the $18$-configuration target-strength matrix across DanceOPD, DiffusionOPD, and Flow-OPD plus per-channel 8-bit quantized references, which match float16 within $0.7\%$ on Editing Mean.
In Figure~\ref{fig:extension-judge-ranks}, GPT-5.5 and InternVL2.5-8B reach $0.928$ Pearson correlation on Editing Structure and $0.921$ on Editing Mean with $85.9\%$ pairwise winner agreement across all $32$ distillation and ablation endpoints ($84.1\%$ pairwise agreement and $r=0.92$ OLS correlation on the six-endpoint $96$-condition audit across the three diagnostic editing routes in Figure~\ref{fig:extension-judge-ranks}(b)), with PrefixOPD ranking first under both judges.

\begin{table}[htbp]
\centering
\includegraphics[width=\textwidth]{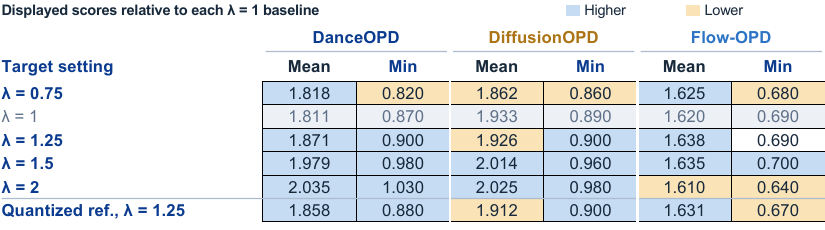}
\caption{\textbf{Equal-data target-strength matrix across base hosts.} Extrapolation improves DanceOPD and DiffusionOPD across target strengths while leaving Flow-OPD nearly unchanged.}
\label{tab:lambda-complete}
\end{table}

\begin{figure}[htbp]
\centering
\includegraphics[width=\textwidth]{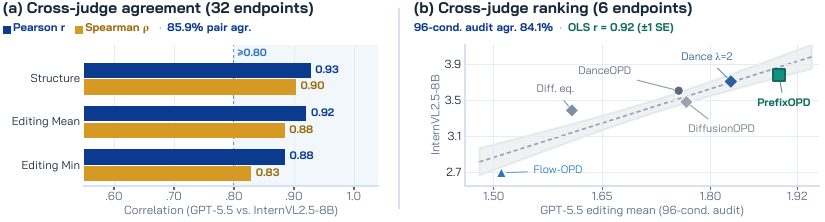}
\vspace{-0.38cm}
\caption{\textbf{Cross-judge agreement audit.} (a) Cross-judge Pearson and Spearman correlations between GPT-5.5 and InternVL2.5-8B across all $32$ evaluated endpoints ($85.9\%$ pairwise agreement), and (b) six-endpoint $96$-condition audit with OLS linear fit ($\pm 1$ SE band, $r=0.92$, $84.1\%$ pairwise agreement).}
\label{fig:extension-judge-ranks}
\end{figure}

\endgroup

\clearpage
\bibliography{references}
\bibliographystyle{report-references}

\end{document}